%% file: main.tex
\pdfoutput=1
\ifdefined\pdfsuppressptexinfo\pdfsuppressptexinfo=-1\fi

\documentclass[10pt,twocolumn,letterpaper]{article}

\usepackage[pagenumbers]{cvpr}

\input{preamble}

\definecolor{cvprblue}{rgb}{0.21,0.49,0.74}
\usepackage[pagebackref,breaklinks,colorlinks,allcolors=cvprblue]{hyperref}

\def\confName{CVPR}
\def\confYear{2027}

\title{LoRAccord: Label-Free Model Merging of LoRA Adapters for\\Vision-Language Models Trained on Data That Cannot Be Shared}

\author{%
Weiyue Li$^{1}$\quad
Yiyang Zhang$^{1,2}$\quad
Gabriel Azpitarte$^{1,3}$\quad
Dachuan Zhao$^{1}$\quad
Yiwen Li$^{1}$\\
Alexandre Molohio$^{1,4}$\quad
Minda Zhao$^{1}$\quad
Yan Luo$^{1\dagger}$\quad
Mengyu Wang$^{1\dagger}$\\[0.4em]
$^{1}$Harvard University\qquad
$^{2}$Washington University in St.\ Louis\\
$^{3}$\'Ecole Polytechnique\qquad
$^{4}$Sorbonne Universit\'e
}

\begin{document}
\maketitle
{\let\thefootnote\relax
\footnotetext{$^{\dagger}$Contributed equally as co-senior authors.}
\footnotetext{Code: \href{https://github.com/Harvard-AI-and-Robotics-Lab/LoRAccord}{Harvard-AI-and-Robotics-Lab/LoRAccord}}}
\input{sec/0_abstract}
\input{sec/1_intro}
\input{sec/2_related}
\input{sec/3_method}
\input{sec/4_experiments}
\input{sec/5_conclusion}

{
    \small
    \bibliographystyle{ieeenat_fullname}
    \bibliography{main}
}

\input{sec/X_suppl}

\end{document}

%% file: preamble.tex
\usepackage{microtype}

\usepackage{multirow}
\usepackage{colortbl}
\usepackage{pifont}
\usepackage{amsthm}
\usepackage{algorithm}
\usepackage{algpseudocode}
\usepackage{tcolorbox}

\definecolor{ours}{RGB}{232,240,250}
\newcommand{\oursrow}{\rowcolor{ours}}
\newcommand{\cmark}{\ding{51}}
\newcommand{\xmark}{\ding{55}}
\newcommand{\tsub}[1]{\textsuperscript{#1}}
\newcommand{\Gt}{\tilde{G}}
\newcommand{\method}{LoRAccord}

\newtheorem{observation}{Observation}

%% file: sec/0_abstract.tex
\begin{abstract}
Sharing model weights is much easier than sharing data. Hospitals and other data holders fine-tune LoRA adapters of one open vision or vision-language model on data that cannot leave them, and want one adapter that does what all of theirs do, built without moving data or labels and served without a router. RegMean merges such adapters in closed form from statistics of each layer's inputs, but fixes the scale of the merged update at $1/K$ for $K$ adapters. The best scale moves from merge to merge: below $1/K$ on CLIP, above it on the medical models. \method{} is a training-free, label-free model merging method. It solves the regression once from base-model statistics, picks the scale of each merge by its agreement with each adapter on unlabeled inputs, and returns one exact LoRA. On all seven pools of our ablation, this label-free pick is the scale that labeled validation data would select. On two CLIP backbones, three medical VLMs and Gemma-4-31B, \method{} has a higher average than every baseline we compare, up to 8.8 points above RegMean. It also leads on four language-model pools, and merges ten dermoscopy sites into one adapter 8.0 points above the best site.
\end{abstract}

%% file: sec/1_intro.tex
\section{Introduction}
\label{sec:intro}

\begin{figure*}[t]
\centering
\includegraphics[width=\textwidth]{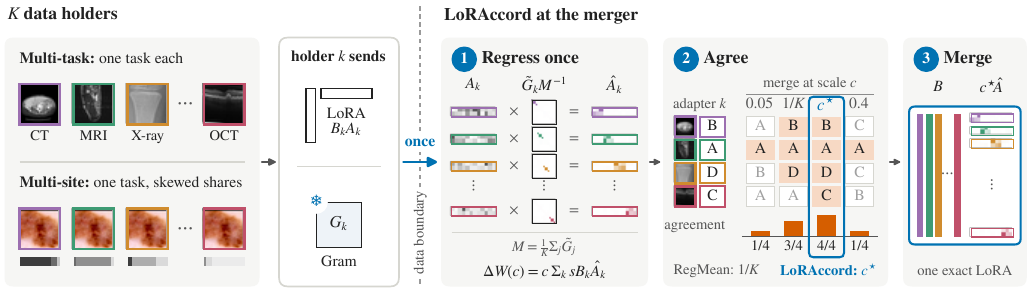}
\caption{\textbf{\method{}: one message per holder, one scale picked without labels, one exact LoRA.} Left: each of $K$ holders has a LoRA trained on its own task (multi-task) or on its own share of one task (multi-site; bars: image sources). It sends the LoRA $B_kA_k$ and the Gram $G_k$ of every adapted layer, recorded once by the frozen base model (\ding{100}) on 64 unlabeled inputs. Right: \textbf{1}~Regress once (RegMean in closed form; $\tilde G_k M^{-1}$ weights adapter $k$ toward the input directions its task uses), \textbf{2}~Agree (shaded: the merge at scale $c$ gives adapter $k$'s answer; RegMean fixes $c = 1/K$, \method{} keeps the $c^\star$ that agrees most), \textbf{3}~Merge (one exact LoRA of factor width $Kr$, no router). Heatmaps and answers are illustrative.}
\label{fig:method}
\end{figure*}

Open vision and vision-language models~\cite{radford2021clip,chen2024internvl,bai2025qwen3vl,hu2024bliva} are now adapted with LoRA~\cite{hu2022lora} by many parties, each on data that cannot leave it. Hospitals fine-tune a medical vision-language model on their own imaging studies, one adapter per modality. Within one organization, data-residency rules can keep each region's images with a regional team that trains the same model for the same task. The next step is one adapter that does what all of them do: served in one backbone pass with no router~\cite{chen2026concur}, and built without moving data or labels. We study two settings (\cref{fig:method}, left). In \emph{multi-task} merging the adapters cover different tasks, such as imaging modalities or classification datasets, and the merge should keep every capability. In \emph{multi-site} merging every adapter covers the same task, trained on one site's skewed share of the data, and the merge should beat every site.

\begin{figure}[t]
\centering
\includegraphics[width=\linewidth]{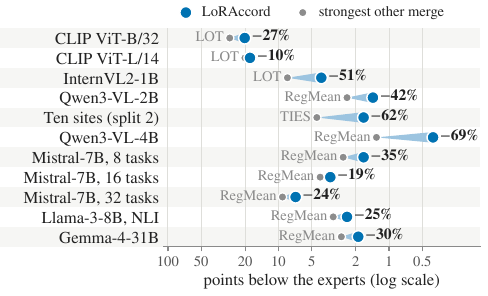}
\caption{\textbf{\method{} beats the strongest other merge on every pool.} Points below the experts (ten sites: below the pooled LoRA) on a log scale; rival: the strongest other merge in \cref{tab:clip,tab:medvlm,tab:gemma,tab:llm,tab:sites}; labels: share of the rival's gap that \method{} closes.}
\label{fig:teaser}
\end{figure}

Existing merges trade accuracy against what they need. Weights-only merges such as task arithmetic~\cite{ilharco2023editing} and TIES~\cite{yadav2023ties} need only the adapters but lose much of the experts' accuracy on dissimilar tasks. AdaMerging~\cite{yang2024adamerging}, NSC~\cite{lee2026nsc} and Expert Merging~\cite{zhang2025expertmerging} learn merge coefficients by gradient descent on unlabeled inputs. RegMean~\cite{jin2023regmean} needs neither labels nor gradients: it solves a least-squares problem per layer in closed form from the Gram matrices of each expert's layer inputs, and gives every regressed update the same coefficient, $1/K$ for $K$ adapters.

Our main observation is that the best value of this one coefficient changes from merge to merge, and that fixing it at $1/K$ costs up to 4.1 points (\cref{tab:ablate}). The value $1/K$ is the best scale for each layer taken alone (\cref{sec:scale}), but the merged network is scored end to end, and its best scale need not be $1/K$. Measured on five vision and vision-language models, the best scale lies below $1/K$ on CLIP and above it on the medical models (\cref{fig:scale}). No fixed value serves every merge: 0.2, the best scale on Qwen3-VL-2B and InternVL2-1B, loses 26 points on CLIP ViT-B/32. The cost of $1/K$ also grows as holders join, because $1/K$ shrinks every update while the best scale settles at 0.2 to 0.25: on InternVL2-1B the gap reaches 3.9 points at $K = 8$ (\cref{fig:ksweep}). The best scale can still be found without labels. The merged model's agreement with each adapter on unlabeled inputs peaks at the most accurate scale, and it ranks merges of different methods as accuracy does (\cref{fig:agree}). Entropy, which AdaMerging minimizes, does not: the scales it picks are 13.2 points less accurate on ViT-B/32 and 9.7 on InternVL2-1B (\cref{tab:scorers}).

We turn this observation into \method{} (\cref{fig:method}). Each data holder sends its adapter and, once, the input Gram of every adapted layer, recorded by the frozen base model on a few dozen unlabeled inputs. The merger solves the regression in closed form, picks one shared scale for this merge by agreement with every adapter on unlabeled inputs, and returns one exact LoRA. Our contributions are:
\begin{itemize}
\item We show that the best scale of RegMean's merged update lies on either side of $1/K$ and moves with the adapters being merged, and that the merged model's agreement with the adapters on unlabeled inputs finds it: on all seven pools of \cref{tab:ablate}, the label-free pick equals the labeled pick (\cref{sec:scale,sec:res-agree}).
\item We propose \method{}, which regresses on base-model statistics, picks the scale for each merge without labels or gradients, and returns one exact LoRA. It needs one message per holder, plus one scoring round if only the holders have unlabeled inputs, and only the adapters if the merger has them.
\item On CLIP ViT-B/32 and ViT-L/14, three medical VLMs and Gemma-4-31B, \method{} has a higher average than every baseline we compare, weights-only, activation-aware or learned. From ten sites with skewed shares of one task it builds one adapter that beats every site (\cref{sec:experiments}, \cref{fig:teaser}). The same merge also has the best average on four pools of language-model adapters of Llama-3-8B and Mistral-7B (\cref{sec:res-llm}).
\end{itemize}

%% file: sec/2_related.tex
\section{Related work}
\label{sec:related}

\paragraph{Model merging.} Weights-only merges average or edit task vectors~\cite{wortsman2022soups,ilharco2023editing,yadav2023ties,yu2024dare}, and spectral methods align, compress or flatten the singular subspaces of the updates: TSV~\cite{gargiulo2025tsv} and isotropic merging~\cite{marczak2025isoc} for full fine-tuning, KnOTS~\cite{stoica2025knots} and Core Space~\cite{panariello2025core} for LoRA. These use the weights alone, and several tune their scale on labeled validation data. Data-aware merges use statistics of each expert on its own data: Fisher merging weights parameters by their Fisher information~\cite{matena2022fisher}, RegMean~\cite{jin2023regmean} regresses on layer inputs, and MaTS~\cite{tam2024mats} casts both as one linear system. LOT Merging~\cite{sun2025lot} solves the same regression on the experts' features and scales it by one global coefficient, set per backbone; its value 1 is $c = 1/K$. LoRM~\cite{salami2025lorm} solves it for LoRA factors, one at a time in closed form. IterIS~\cite{chen2025iteris}, RegMean++~\cite{nguyen2026regmeanpp} and Chain of Merges~\cite{buzzega2026chain} recompute the inputs on the current merge, so their statistics change with every merge, over several rounds when the inputs stay with the holders; \method{}'s need no adapter and are sent once.

\paragraph{Learning merge coefficients.} AdaMerging~\cite{yang2024adamerging} learns task- or layer-wise coefficients by minimizing prediction entropy on unlabeled inputs, and NSC~\cite{lee2026nsc} learns coefficients for LoRA merges from a label-free objective on the adapters' down-projections. OptMerge~\cite{wei2026optmerge} instead optimizes the merged update itself under a data-free loss on the task vectors, FroM~\cite{li2025from} fits a merge data-free under a Frobenius-norm objective, and MetaGPT~\cite{zhou2024metagpt} sets task-arithmetic coefficients from the weights alone. AdaMMS~\cite{du2025adamms} picks one interpolation coefficient for merging multimodal LLMs without labels or gradients, from the consistency of the merged model's outputs across candidate coefficients, Expert Merging~\cite{zhang2025expertmerging} trains layer-wise coefficients that align the merge with the experts on unlabeled inputs, and Representation Surgery~\cite{yang2024surgery} trains a module per task that does so for the merged representations. LoraHub~\cite{huang2024lorahub} searches LoRA weights without gradients on a few labeled examples. Evolutionary~\cite{akiba2025evo} and Bayesian~\cite{agarwal2026mobo} searches score many candidate merges.

\paragraph{Federated and one-shot learning.} Federated averaging~\cite{mcmahan2017fedavg} and its LoRA variants~\cite{zhang2024fedit,sun2024ffalora,singhal2025fedex} take many rounds. One-shot federated learning aggregates once: FedFisher sends each local model with its Fisher information~\cite{jhunjhunwala2024fedfisher}, one round of aggregation can match many when clients fine-tune a large foundation model~\cite{wang2024singleround}, and task arithmetic can itself be read as one-shot FedAvg~\cite{tao2025oneshot}. Fed3R~\cite{fani2024fed3r} sends feature second moments once and solves a classifier in closed form. \method{} also takes one round; it merges adapters that already exist, trained however their owners chose, by regression on base-model statistics.

\paragraph{What is new.} \looseness=-1 New in \method{} is the finding that the best scale of RegMean's merged update lies on either side of $1/K$ and moves with the backbone and the adapters, so no fixed value serves every merge, and that one scalar searched by agreement finds it and improves RegMean on the experts' own statistics on every pool. Unlike LOT's per-backbone coefficient and AdaMMS's consistency among candidates, \method{} picks per merge by agreement with each task's own adapter on its inputs. Statistics of the frozen base model add reuse: the message depends on no adapter, serves every later merge and can be computed by the merger, and at RegMean's scale and shrinkage they match or beat the experts' own. \method{} builds on RegMean's regression~\cite{jin2023regmean}, one scalar on a merged update as in task arithmetic~\cite{ilharco2023editing}, coefficients chosen without labels~\cite{yang2024adamerging,lee2026nsc} and exact factor stacking~\cite{wang2024flora}.

%% file: sec/3_method.tex
\section{Method}
\label{sec:method}

\subsection{Setting}
\label{sec:setting}

$K$ data holders share a public base model with weights $W_0^\ell$ at each adapted layer $\ell$. Holder $k$ has fine-tuned a LoRA adapter $\Delta W_k^\ell = s\,B_k^\ell A_k^\ell$, with $B_k^\ell \in \mathbb{R}^{d_{\text{out}} \times r}$, $A_k^\ell \in \mathbb{R}^{r \times d_{\text{in}}}$ and $s = \alpha / r$, on labeled data that stays with it, and it can run the base model on a few of its own unlabeled inputs $X_k$. A merger receives one message from each holder and returns one adapter $\Delta W^\ell$ per layer, served as $W_0^\ell + \Delta W^\ell$ for every task, with no router and no further training. The merger holds no labeled data. It may hold a few unlabeled inputs $U_k$ for each task, such as inputs a holder agreed to release; otherwise each holder scores the merger's candidates on its own inputs (\cref{sec:pick}).

\paragraph{What each party sees.} Labels and training data stay with their holder. The merger is trusted with the adapters and the Grams below, second moments of base-model layer inputs, or computes the Grams from released inputs. Holders that score the candidates receive the merge's stacked factors, as every holder does at the end.

\subsection{RegMean with base-model Grams}
\label{sec:message}

Fix one adapted layer and drop $\ell$. Holder $k$ runs the frozen base model on $X_k$ and records the per-token mean Gram of the layer's input,
\begin{equation}
G_k = \tfrac{1}{T_k} \textstyle\sum_{t=1}^{T_k} x_t x_t^\top \in \mathbb{R}^{d_{\text{in}} \times d_{\text{in}}},
\end{equation}
over all $T_k$ tokens of the input, image and prompt, without an answer; projections that share an input share its Gram. With these inputs stacked as the rows of $Z_k \in \mathbb{R}^{T_k \times d_{\text{in}}}$, so that $G_k = Z_k^\top Z_k / T_k$, RegMean~\cite{jin2023regmean} finds the weight that best reproduces every expert on its own inputs, $\min_W \sum_k \frac{1}{T_k}\lVert Z_k W^\top - Z_k W_k^\top \rVert_F^2$, solved by $W^\star = (\sum_k W_k G_k)(\sum_k G_k)^{-1}$. With $W_k = W_0 + \Delta W_k$ the base weight passes through, so merging the deltas is exact RegMean on the full weights. Following~\cite{jin2023regmean} we shrink the off-diagonal entries, $\Gt_k = 0.9\,G_k + 0.1\,\mathrm{diag}(G_k)$, and with $M = \frac{1}{K}\sum_j \Gt_j$,
\begin{equation}
\Delta W_{\text{RegMean}} = \frac{1}{K} \sum_{k=1}^{K} \Delta W_k \, \Gt_k M^{-1}.
\label{eq:regmean}
\end{equation}
We solve with $M + \epsilon I$ in place of $M$, with $\epsilon$ equal to $10^{-6}$ times the mean diagonal of $M$; \cref{eq:regmean} is then exactly the minimizer of $\sum_k \lVert (\Delta W - \Delta W_k)\Gt_k^{1/2} \rVert_F^2 + K\epsilon \lVert \Delta W \rVert_F^2$. RegMean records $G_k$ inside expert $k$'s fine-tuned model, so its Gram at layer $\ell$ depends on the expert's adapters below $\ell$. \method{} records it in the base model, so the message depends on no adapter: it serves every later merge of any subset, and a merger with unlabeled inputs can compute it alone. The merger needs only $A_k\Gt_k$, 15\,MB per holder on Qwen3-VL-2B, and $\sum_j \Gt_j$, which a secure sum~\cite{bonawitz2017secure} delivers without exposing any one Gram.

\subsection{The best scale moves}
\label{sec:scale}

We separate the scale from the regression, as in~\cite{sun2025lot}, with one scalar $c$ shared by every adapted layer,
\begin{equation}
\Delta W(c) = c \sum_{k=1}^{K} \Delta W_k \, \Gt_k M^{-1},
\label{eq:ours}
\end{equation}
so that $c = 1/K$ is RegMean. For any Grams, $c = 1/K$ minimizes the objective of \cref{sec:message}: it is the scale at which each layer, taken alone, best reproduces every expert on that expert's inputs. Two limiting cases show what it reduces to.

\begin{observation}
\label{obs:bracket}
(i) If $\Gt_k = \Gt$ for every $k$, with $\Gt$ nonsingular, then $\Delta W(c) = c\sum_k \Delta W_k$, task arithmetic with coefficient $c$ (exactly for $\epsilon = 0$). (ii) If the raw Grams $G_k$ have mutually orthogonal column spaces $S_k$, with no shrinkage, $\epsilon = 0$ and $M^{-1}$ the pseudo-inverse, then $\Delta W(c)\,x = cK\,\Delta W_k\,x$ for $x \in S_k$, so $c = 1/K$ reproduces each adapter exactly on its own inputs.
\end{observation}

\begin{figure*}[t]
\centering
\includegraphics[width=\textwidth]{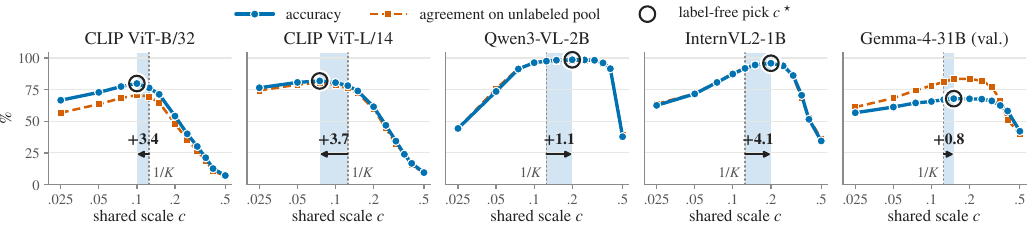}
\caption{\textbf{The best scale lies on either side of $1/K$.} Accuracy (blue) and agreement with the adapters on unlabeled inputs (orange) of RegMean on base-model Grams, against the scale $c$. Circle: the label-free pick $c^\star$, the most accurate scale in every panel; dotted: $1/K$; arrow: from $1/K$ to $c^\star$, with the points of accuracy gained. Gemma-4-31B: raw validation accuracy; others: normalized test accuracy.}
\label{fig:scale}
\end{figure*}

The proof is in Supp.~C. In these two limits, $c = 1/K$ averages the updates where all holders share one Gram and applies each update in full on inputs that only its own holder's Gram spans. The merged model, however, is scored end to end. Each merged layer receives the outputs of the merged layers below it, while its Gram was recorded on base-model inputs, and accuracy counts final answers, not each layer's squared error. The layerwise optimum $1/K$ therefore need not be the most accurate scale, and we treat $c$ as a calibration of the whole merged model. \Cref{fig:scale} measures the best scale directly. On CLIP it lies below $1/K$, on the medical models above it, with or without shrinkage. \method{} therefore picks $c$ for each merge by agreement.

\subsection{Picking the scale without labels}
\label{sec:pick}

The merger picks $c$ from a grid $\mathcal{C}$ of scales (twelve from 0.025 to 0.5 on the pools of \cref{fig:scale}) by the merged model's agreement with each adapter on unlabeled inputs,
\begin{equation}
c^\star = \arg\max_{c \in \mathcal{C}} \frac{1}{K} \sum_{k=1}^{K} \mathrm{agree}\big(f_{\Delta W(c)}, f_k; U_k\big),
\label{eq:pick}
\end{equation}
\looseness=-1 where $\mathrm{agree}$ is the fraction of inputs in $U_k$ on which the merged model and adapter $k$ give the same answer (the same class, option or extracted answer under each benchmark's scoring rule). Let $a_k > 0$ be adapter $k$'s accuracy on $U_k$. On a task scored against one correct answer, a merge is right wherever it agrees with a right adapter, so its accuracy on $U_k$ differs from its agreement by at most $1 - a_k$, and the scale that agrees most is within $\frac{3}{K}\sum_k (1 - a_k)/a_k$ of the most accurate scale on the pool, in normalized accuracy. The bound is small where the adapters are accurate on their own inputs: with the experts' test accuracies plugged in for $a_k$, it is at most 1.5 points on the medical VLMs (Supp.~C). On every pool we also measure the pick against the most accurate scale on test (\cref{sec:res-ablate}). The pick needs only forward passes of the candidates and is repeated for every merge (\cref{fig:ksweep}). Agreement between models on unlabeled data is an established proxy for accuracy~\cite{jiang2022disagreement,baek2022agreement}; \method{} measures it between the merge and each task's own adapter, on that task's inputs. On the sites, $\mathcal{C}$ also holds FedAvg size weights~\cite{mcmahan2017fedavg}, $cKn_k/N$ in place of $c$ for a site with $n_k$ of the $N$ training inputs, and the plain sum of the adapters (Supp.~A).

\paragraph{Without inputs at the merger.} The merger sends the stacked factors $(B, \hat A)$ to every holder once, and each holder scores all $|\mathcal{C}|$ scales on its own inputs and returns $|\mathcal{C}|$ numbers, a second and last exchange.

\paragraph{The merged adapter.} Stacking the factors, as FLoRA does for federated LoRA~\cite{wang2024flora}, makes \cref{eq:ours} an exact LoRA: $\Delta W(c) = \sum_k (sB_k)(cA_k\Gt_k M^{-1}) = BA$, with $B = [sB_1 \cdots sB_K] \in \mathbb{R}^{d_{\text{out}} \times Kr}$ and $A$ stacking the rows $cA_k\Gt_kM^{-1}$. Its factor width is $Kr$, 128 for eight rank-16 adapters, and it is served as one adapter in one backbone pass or folded into $W_0$ at no serving cost (\cref{alg:merge}).

\begin{algorithm}[t]
\caption{\method{}. $\mathcal{C}$: candidate scales (\cref{sec:pick}). Without unlabeled inputs at the merger, the holders run lines 8--9.}
\label{alg:merge}
\begin{algorithmic}[1]
\State \textbf{Holder $k$} (once): run the frozen base model on $X_k$; send $(B_k^\ell, A_k^\ell, G_k^\ell)$ for every layer $\ell$
\For{every layer $\ell$} \Comment{merger; drop $\ell$}
  \State $\Gt_k \gets 0.9\,G_k + 0.1\,\mathrm{diag}(G_k)$ for all $k$
  \State $M \gets \frac{1}{K}\sum_j \Gt_j + \epsilon I$
  \State $\hat A_k \gets A_k \Gt_k M^{-1}$ by a solve, for all $k$
\EndFor
\State $B \gets [sB_1 \cdots sB_K]$, $\hat A \gets [\hat A_1; \ldots; \hat A_K]$
\For{$c \in \mathcal{C}$} \Comment{$f_c$: the merge $(B, c\hat A)$}
  \State score $\frac{1}{K}\sum_k \mathrm{agree}(f_c, f_k; U_k)$
\EndFor
\State \Return $(B, c\hat A)$ at the best $c$
\end{algorithmic}
\end{algorithm}

%% file: sec/4_experiments.tex
\section{Experiments}
\label{sec:experiments}

\input{tables/clip}

\begin{figure*}[t]
\centering
\includegraphics[width=\textwidth]{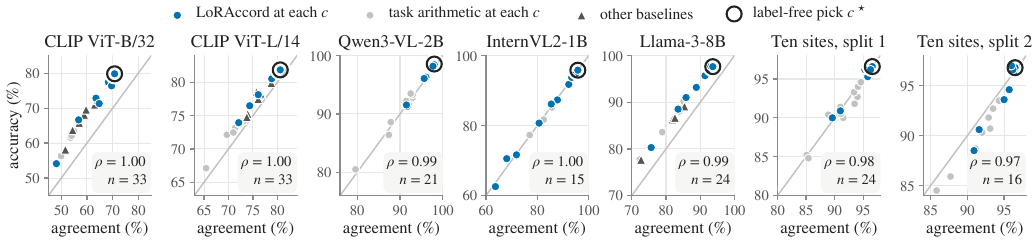}
\caption{\textbf{Agreement ranks merges as accuracy does.} Each point is one merge, scored by agreement with the adapters on the unlabeled pool and by test accuracy: \method{} (blue) and task arithmetic (gray) at each scale, and on CLIP and Llama-3-8B the other merges of the tables (triangles). Circle: the label-free pick. $\rho$: Spearman over the $n$ merges; gray line: equal agreement and accuracy.}
\label{fig:agree}
\end{figure*}

\input{tables/medical}
\input{tables/sites_ablate}
\input{tables/cost}

\subsection{Setup}
\label{sec:setup}

\paragraph{Benchmarks.} \emph{CLIP:} the eight rank-16 LoRA experts released with KnOTS~\cite{stoica2025knots} on the vision towers of CLIP~\cite{radford2021clip} ViT-B/32 and ViT-L/14~\cite{dosovitskiy2021vit}, for SUN397~\cite{xiao2010sun}, Stanford Cars~\cite{krause2013cars}, RESISC45~\cite{cheng2017resisc}, EuroSAT~\cite{helber2019eurosat}, SVHN~\cite{netzer2011svhn}, GTSRB~\cite{stallkamp2011gtsrb}, MNIST~\cite{lecun1998mnist} and DTD~\cite{cimpoi2014dtd}, scored with zero-shot text heads as in KnOTS. \emph{Medical VLMs:} MergeMedBench~\cite{mou2026mergemedbench} releases one LoRA expert on the language model per imaging modality (CT, MRI, X-ray, ultrasound, dermoscopy, fundus, OCT, microscopy) for Qwen3-VL-2B~\cite{bai2025qwen3vl} and InternVL2-1B~\cite{chen2024internvl,chen2024internvl15}, trained on multiple-choice questions from OmniMedVQA~\cite{hu2024omnimedvqa}; we train the same eight for Qwen3-VL-4B. \emph{Gemma-4-31B:} eight LoRA adapters of Gemma-4-31B~\cite{gemma2026gemma4}, trained with the recipe of the fine-tuned comparators of~\cite{zhao2026rgicl}, for four medical image classification and four medical VQA tasks. \emph{Ten sites:} one task, dermoscopy VQA on Qwen3-VL-2B. We split OmniMedVQA's dermoscopy training questions by image over ten simulated sites with Dirichlet(0.1) proportions over the seven image sources~\cite{hsu2019noniid}, train one LoRA per site, and test on MergeMedBench's dermoscopy test set, for two splits. Supp.~A gives sizes, splits and training details.

\paragraph{Metric.} In the multi-task setting, normalized accuracy: each task's merged accuracy divided by its own expert's, averaged over tasks~\cite{stoica2025knots,mou2026mergemedbench}. In the multi-site setting, raw accuracy on the shared test set. Differences carry paired bootstrap standard errors over test items, and \method{}'s lead over RegMean exceeds twice its SE on every vision, vision-language and NLI pool (Supp.~D).

\paragraph{Baselines.} \emph{Weights only:} weight averaging, task arithmetic (TA)~\cite{ilharco2023editing} and TIES~\cite{yadav2023ties}; TSV-M~\cite{gargiulo2025tsv} and KnOTS-TIES~\cite{stoica2025knots} on CLIP; winner-take-all (WTA)~\cite{mou2026mergemedbench} on the medical models. \emph{Activation-aware:} RegMean~\cite{jin2023regmean}, with each expert's own Grams and $c = 1/K$, and on CLIP and the medical VLMs LOT~\cite{sun2025lot}, the same regression without shrinkage solved by the pseudo-inverse, at $\lambda = 1$, which is $c = 1/K$. \emph{Learned:} AdaMerging~\cite{yang2024adamerging}, layer-wise, NSC~\cite{lee2026nsc} and, on the medical VLMs, Expert Merging~\cite{zhang2025expertmerging}, at their published settings and on the same unlabeled pool as \method{}'s pick. On CLIP the weights-only merges and NSC's initial coefficient are tuned on labeled validation images, as in KnOTS; RegMean and LOT keep $c = 1/K$, and RegMean and \method{} share the off-diagonal shrinkage 0.9, tuned for neither. On the medical models TA's coefficient is picked by \cref{eq:pick} and TIES uses MergeMedBench's setting. Supp.~B gives our readings where papers leave details open.

\paragraph{Implementation.} Each holder records its Grams on 64 unlabeled inputs per task (32 per site), and RegMean the experts' Grams on the same inputs. Ties in agreement go to the smaller $c$. Generation is greedy, and an answer outside the required format counts as wrong.

\subsection{\method{} leads on both CLIP backbones}
\label{sec:res-clip}

\method{} has the best average of the merges in \cref{tab:clip} on both backbones, 8.8 points above RegMean on ViT-B/32 (SE 0.29). Weights-only merges and the learned AdaMerging and NSC keep SUN397, where the zero-shot model is already close to the expert, and lose most of the experts' gain on SVHN and GTSRB. RegMean recovers much of that gain on ViT-B/32 but gives up part of SUN397 and EuroSAT, which the picked scale largely keeps.

\subsection{Best average on every medical VLM}
\label{sec:res-medvlm}

\method{} has the best average on all three medical VLMs (\cref{tab:medvlm}). Its lead is largest on InternVL2-1B, whose experts share an initialization and overlap most: 4.7 points above RegMean (SE 0.53). Ultrasound on InternVL2-1B shows where the gain comes from: every weights-only and learned merge keeps at most two thirds of the expert's accuracy, regression at $1/K$ recovers most of the gap, and the picked scale nearly all of it. The learned merges stay below RegMean on all three models, Expert Merging included, although it trains its coefficients to match each expert on the same unlabeled pool. \method{} also leads on two more pools: medical plus spatial adapters on Qwen3-VL-2B and natural-image VQA adapters on Qwen2-VL-7B (Supp.~E, F).

\subsection{Scaling to a 31B model}
\label{sec:res-gemma}

On the eight Gemma-4-31B adapters, which cover image classification and open-ended VQA and also adapt the vision projector, \method{} has the best average (\cref{tab:gemma}), 7.6 points above AdaMerging (SE 0.52), with its scale picked on unlabeled validation questions (\cref{fig:scale}).

\subsection{The same merge works on language models}
\label{sec:res-llm}

Nothing in \method{} is specific to images. On the language-model adapters of \cref{tab:llm} it runs unchanged: base-model Grams on 64 unlabeled prompts per task, and the scale picked by agreement with each adapter on unlabeled prompts, which on generation is the ROUGE-L between the merge's output and the adapter's. It has the best average on all four pools, and its lead over RegMean is largest with 32 generation adapters, 2.2 points. AdaMerging and NSC learn on the same prompts and lose ground as $K$ grows.

\subsection{One adapter beats every site}
\label{sec:res-sites}

\method{} turns ten site adapters into one adapter that beats every site (\cref{tab:sites}). We chose the pick rule on split~1: each site's agreement is scored on its own pool questions, and the pick also chooses equal or size weights; split~2 is held out. On the held-out split it leads the best site by 8.0 points and the best other merge by 2.8, and comes within two points of a LoRA trained on pooled data, which the sites cannot share.

\subsection{Where the gain comes from}
\label{sec:res-ablate}

\Cref{tab:ablate} separates the two changes \method{} makes to RegMean and asks whether agreement finds the scale. At RegMean's $1/K$ and shrinkage, base-model Grams match or beat the experts' own on every pool, most on ViT-B/32. \method{}'s pick then lifts accuracy above $1/K$ on every pool, as it does for RegMean's own Grams (Supp.~C), and base-model Grams add a message that depends on no adapter. On every pool the label-free pick for base-model Grams equals both the labeled pick and the most accurate scale scored on test (three scales on Gemma-4-31B). Redrawing the pool changes little: over 2,000 bootstrap draws of the pool items, the pick's mean loss against the best candidate on test is at most 0.4 points on both CLIP backbones, both medical VLMs and both site splits (Supp.~G). With Grams the merger computes on its pool inputs, \method{} stays within half a point of the holders' Grams, so holders that trust the merger with unlabeled inputs send only their adapters. Few unlabeled inputs suffice: Grams level off at 32 calibration questions per task on InternVL2-1B, and 50 pool questions per task pick the same scale as 250 (Supp.~G).

\subsection{Agreement finds the most accurate merge}
\label{sec:res-agree}

Agreement on unlabeled inputs ranks merges as accuracy does, across methods and across scales (\cref{fig:agree}). In every pool Spearman's $\rho$ is at least 0.97, and the merge that agrees most with the adapters is the most accurate or within noise of it. When LOT~\cite{sun2025lot}, isotropic~\cite{marczak2025isoc} and Core Space~\cite{panariello2025core} merges join the candidates on CLIP ViT-B/32, the merge that agrees most is still the most accurate (Supp.~B). On the same candidates and pools, agreement is the only label-free scorer that loses nothing on all four (\cref{tab:scorers}); AdaMMS's consistency~\cite{du2025adamms}, entropy and KL from each adapter lose up to 26 points on CLIP. Matching the adapters' features instead of their answers favors $1/8$ or 0.15 on every pool, near the layerwise optimum, and loses up to 8.5 points. A scale computed from the adapters and Grams alone, fit on 76 other merges, loses up to 15.5 points on the Qwen3-VL-2B merges; the pick loses none (Supp.~C).

\input{tables/scorers}

\subsection{The gain grows with the number of adapters}
\label{sec:res-ksweep}

The picked scale falls more slowly than $1/K$ (\cref{fig:ksweep}), and its median settles at 0.2 to 0.25 from five adapters on. Accuracy at $1/K$ falls with $K$, while at the picked scale it stays near the experts, so the pick's gain reaches 3.9 points at $K = 8$.

\subsection{No labels, no gradients, minutes on one GPU}
\label{sec:res-cost}

\method{} needs no labels or gradients and runs faster than the learned merges on all four models of \cref{tab:cost}, 2 to 32 times faster than AdaMerging (3.8 s against 123 s on ViT-B/32).

%% file: tables/clip.tex
\begin{table*}[t]
\centering\scriptsize\setlength{\tabcolsep}{2.6pt}
\caption{\textbf{\method{} has the best average on both CLIP backbones.} Top: accuracy (\%) of the zero-shot model and the experts. Bottom: merges, normalized to each task's expert (\%). Bold: best merge; underlined: second best. On ViT-L/14 task arithmetic picks $c = 1/8$, which is weight averaging.}
\label{tab:clip}
\begin{tabular}{l|rrrrrrrr|r!{\vrule width 0.8pt}rrrrrrrr|r}
\toprule
 & \multicolumn{9}{c!{\vrule width 0.8pt}}{\textbf{CLIP ViT-B/32}} & \multicolumn{9}{c}{\textbf{CLIP ViT-L/14}} \\
\cmidrule(lr){2-10}\cmidrule(lr){11-19}
Method & SUN & Cars & RES & Euro & SVHN & GTSRB & MNIST & DTD & Avg & SUN & Cars & RES & Euro & SVHN & GTSRB & MNIST & DTD & Avg \\
\midrule
Zero-shot & 63.2 & 59.5 & 60.3 & 45.0 & 31.5 & 32.7 & 48.0 & 43.5 & 48.0 & 68.3 & 77.8 & 71.3 & 61.2 & 58.6 & 50.7 & 76.2 & 54.8 & 64.9 \\
Fine-tuned & 64.5 & 74.1 & 88.0 & 99.7 & 96.0 & 92.9 & 99.3 & 57.8 & 84.0 & 79.3 & 99.8 & 95.7 & 99.7 & 97.6 & 97.2 & 99.6 & 70.3 & 92.4 \\
\midrule
Weight averaging & 96.6 & 82.4 & 71.3 & 45.7 & 43.3 & 44.2 & 52.3 & 71.8 & 63.4 & 87.3 & 80.3 & 80.1 & 59.9 & 64.7 & 60.7 & 80.8 & 78.1 & 74.0 \\
TA~\cite{ilharco2023editing} & \underline{97.9} & 81.8 & 71.4 & 49.2 & 39.3 & 41.0 & 52.6 & 73.9 & 63.4 & 87.3 & 80.3 & 80.1 & 59.9 & 64.7 & 60.7 & 80.8 & 78.1 & 74.0 \\
TIES~\cite{yadav2023ties} & 95.9 & 82.9 & 69.6 & 48.2 & 48.0 & 36.4 & 58.5 & 70.3 & 63.7 & 87.7 & 80.0 & 79.3 & 60.8 & 70.0 & 60.6 & 82.5 & 77.4 & 74.8 \\
TSV-M~\cite{gargiulo2025tsv} & 95.6 & \underline{83.6} & \underline{72.5} & 45.2 & 50.3 & 48.9 & 57.7 & 72.6 & 65.8 & 87.9 & 82.1 & 82.2 & 63.2 & 72.0 & 66.9 & 87.5 & 78.0 & 77.5 \\
KnOTS-TIES~\cite{stoica2025knots} & 95.5 & 83.2 & 71.2 & 49.7 & 53.9 & 49.2 & 68.9 & 73.3 & 68.1 & \textbf{88.5} & 82.7 & 82.3 & 60.7 & \underline{76.8} & 68.8 & 88.5 & 78.1 & 78.3 \\
RegMean~\cite{jin2023regmean} & 84.9 & 78.7 & 71.3 & 38.2 & 73.2 & \underline{67.2} & \underline{89.5} & 64.9 & 71.0 & 86.3 & \underline{83.8} & \underline{89.2} & 62.4 & 73.0 & \underline{85.8} & 63.6 & \underline{81.5} & 78.2 \\
LOT~\cite{sun2025lot}, $\lambda = 1$ & 82.0 & 77.5 & 71.2 & 39.1 & \textbf{79.4} & \textbf{71.5} & \textbf{94.7} & 63.8 & \underline{72.4} & 86.1 & 83.6 & \textbf{89.8} & 65.2 & 74.3 & \textbf{87.2} & 70.9 & \textbf{82.2} & \underline{79.9} \\
AdaMerging~\cite{yang2024adamerging} & \textbf{98.0} & 79.9 & 71.3 & \textbf{76.3} & 56.0 & 37.3 & 63.7 & \underline{74.2} & 69.6 & 87.0 & 79.0 & 78.2 & \textbf{74.6} & 67.2 & 66.1 & \textbf{98.1} & 78.2 & 78.6 \\
NSC~\cite{lee2026nsc} & 93.1 & 79.7 & 65.4 & 21.6 & 46.0 & 44.0 & 45.4 & 69.9 & 58.1 & 87.2 & 80.5 & 78.4 & 59.9 & 68.4 & 58.5 & \underline{90.9} & 78.2 & 75.2 \\
\oursrow \textbf{\method{} (Ours)} & 94.2 & \textbf{85.1} & \textbf{76.8} & \underline{75.5} & \underline{76.9} & 65.8 & 87.1 & \textbf{77.0} & \textbf{79.8} & \underline{88.2} & \textbf{84.2} & 85.2 & \underline{71.0} & \textbf{84.1} & 80.0 & 81.2 & 81.0 & \textbf{81.9} \\
\bottomrule
\end{tabular}
\end{table*}

%% file: tables/medical.tex
\begin{table*}[t]
\begin{minipage}[t]{0.485\textwidth}
\input{tables/medvlm}
\end{minipage}\hfill
\begin{minipage}[t]{0.485\textwidth}
\input{tables/gemma}
\vspace{6pt}
\input{tables/llm}
\makeatletter\def\@captype{figure}\makeatother
\centering
\includegraphics[width=0.96\linewidth]{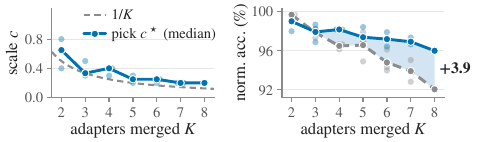}
\caption{\textbf{\method{}'s gain over $1/K$ grows with $K$.} InternVL2-1B, three random $K$-subsets per $K < 8$, 150 test questions per task. Left: picked scale and median. Right: normalized accuracy.}
\label{fig:ksweep}
\end{minipage}
\end{table*}

%% file: tables/medvlm.tex
\centering\scriptsize\setlength{\tabcolsep}{1.9pt}
\caption{\textbf{\method{} has the best average on all three medical VLMs.} MergeMedBench~\cite{mou2026mergemedbench}, 300 test questions per modality. Zero-shot and experts: accuracy (\%); merges: normalized to each expert (\%). Bold: best merge; underlined: second best. \tsub{a}Answer format forced: the option scored highest after \texttt{<answer>}. \tsub{b}Our experts (Supp.~A).}
\label{tab:medvlm}
\begin{tabular}{l|rrrrrrrr|r}
\toprule
Method & CT & MRI & X-ray & US & Derm. & Fund. & OCT & Mic. & Avg \\
\midrule
\multicolumn{10}{c}{\textbf{Qwen3-VL-2B}} \\
\midrule
Zero-shot\tsub{a} & 55.0 & 68.7 & 79.0 & 81.0 & 69.0 & 74.7 & 60.3 & 78.3 & 70.8 \\
Fine-tuned & 100.0 & 99.7 & 98.7 & 100.0 & 98.0 & 100.0 & 100.0 & 99.7 & 99.5 \\
\midrule
Averaging & 75.0 & 91.3 & 90.5 & 90.0 & 84.7 & 94.0 & 79.0 & 86.3 & 86.4 \\
TA & 93.3 & 97.0 & 94.3 & 95.3 & 88.4 & 96.0 & 91.3 & 87.0 & 92.8 \\
TIES & 95.7 & 98.3 & 93.6 & 98.3 & 89.8 & 96.7 & 91.7 & 86.6 & 93.8 \\
WTA & 95.3 & 98.0 & 95.6 & 94.3 & 88.8 & 97.3 & 90.7 & 88.3 & 93.5 \\
RegMean & \underline{99.0} & \underline{99.0} & 97.0 & \underline{99.7} & \underline{94.6} & \underline{98.3} & \underline{96.7} & \underline{96.3} & \underline{97.6} \\
LOT~\cite{sun2025lot}, $\lambda = 1$ & 97.0 & 98.7 & \textbf{97.6} & 99.0 & 93.9 & \underline{98.3} & 95.3 & 94.3 & 96.8 \\
AdaMerging & 94.0 & 97.7 & 92.9 & 99.3 & 86.4 & 96.0 & 93.7 & 85.6 & 93.2 \\
NSC & 90.7 & 94.3 & 95.9 & 99.0 & 88.8 & 96.7 & 91.3 & 90.6 & 93.4 \\
Expert Merging & 92.7 & 97.3 & 94.6 & 95.3 & 89.5 & 95.7 & 92.0 & 86.6 & 93.0 \\
\oursrow \textbf{\method{} (Ours)} & \textbf{99.3} & \textbf{99.7} & \underline{97.3} & \textbf{100.0} & \textbf{96.9} & \textbf{99.3} & \textbf{99.0} & \textbf{97.0} & \textbf{98.6} \\
\midrule
\multicolumn{10}{c}{\textbf{Qwen3-VL-4B\tsub{b}}} \\
\midrule
Zero-shot\tsub{a} & 59.0 & 77.7 & 87.3 & 84.0 & 75.3 & 87.7 & 72.0 & 80.7 & 78.0 \\
Fine-tuned & 99.7 & 99.7 & 98.7 & 100.0 & 99.0 & 99.7 & 100.0 & 99.3 & 99.5 \\
\midrule
Averaging & 95.0 & 96.3 & 96.6 & 95.3 & 87.2 & 94.0 & 96.0 & 93.6 & 94.3 \\
TA & 96.3 & 98.7 & 96.6 & 99.3 & 86.9 & \underline{99.3} & 98.3 & 90.6 & 95.8 \\
TIES & 97.0 & \underline{99.0} & 97.3 & 99.0 & 88.2 & \underline{99.3} & 98.3 & 91.3 & 96.2 \\
WTA & \underline{98.0} & \textbf{99.3} & 96.6 & 96.7 & 89.9 & 98.7 & 98.3 & 93.6 & 96.4 \\
RegMean & \textbf{99.7} & \textbf{99.3} & \underline{97.6} & \underline{99.7} & \underline{95.3} & \textbf{100.0} & \textbf{100.0} & \underline{98.0} & \underline{98.7} \\
LOT~\cite{sun2025lot}, $\lambda = 1$ & \textbf{99.7} & \textbf{99.3} & 97.3 & \underline{99.7} & 93.9 & \textbf{100.0} & \textbf{100.0} & \underline{98.0} & 98.5 \\
AdaMerging & 97.0 & 98.0 & \underline{97.6} & 98.3 & 86.2 & \underline{99.3} & \underline{98.7} & 91.6 & 95.8 \\
NSC & 97.3 & 98.0 & 97.0 & 97.3 & 87.2 & 98.3 & 97.0 & 94.6 & 95.8 \\
Expert Merging & 96.7 & 98.3 & 97.0 & 98.3 & 90.6 & 98.3 & \underline{98.7} & 90.3 & 96.0 \\
\oursrow \textbf{\method{} (Ours)} & \textbf{99.7} & \textbf{99.3} & \textbf{99.0} & \textbf{100.0} & \textbf{99.7} & \textbf{100.0} & \textbf{100.0} & \textbf{99.0} & \textbf{99.6} \\
\midrule
\multicolumn{10}{c}{\textbf{InternVL2-1B}} \\
\midrule
Zero-shot\tsub{a} & 44.0 & 54.3 & 63.7 & 34.3 & 49.3 & 75.0 & 47.7 & 63.3 & 54.0 \\
Fine-tuned & 100.0 & 99.7 & 98.7 & 100.0 & 98.7 & 100.0 & 100.0 & 99.7 & 99.6 \\
\midrule
Averaging & 78.7 & 79.6 & 86.5 & 53.0 & 65.2 & 89.7 & 79.7 & 86.0 & 77.3 \\
TA & 89.3 & 92.6 & 92.2 & 61.3 & 79.4 & 94.0 & 88.0 & 84.3 & 85.2 \\
TIES & 91.7 & \underline{96.3} & \underline{92.6} & 63.3 & \underline{83.4} & 95.3 & 84.3 & 84.9 & 86.5 \\
WTA & 88.0 & 90.6 & 91.9 & 65.3 & 78.4 & 93.0 & 85.0 & 87.0 & 84.9 \\
RegMean & 92.3 & 95.7 & 91.6 & 86.7 & 81.4 & \underline{97.0} & 92.0 & 92.3 & 91.1 \\
LOT~\cite{sun2025lot}, $\lambda = 1$ & \underline{93.3} & \underline{96.3} & 92.2 & \underline{89.0} & 79.7 & \underline{97.0} & \underline{93.0} & \underline{93.0} & \underline{91.7} \\
AdaMerging & 83.3 & 82.3 & 72.6 & 40.7 & 68.6 & 79.3 & 78.0 & 72.6 & 72.2 \\
NSC & 89.3 & 86.6 & 85.1 & 46.0 & 77.7 & 91.3 & 83.3 & 70.2 & 78.7 \\
Expert Merging & 67.0 & 73.6 & 84.5 & 51.0 & 61.5 & 89.7 & 75.7 & 86.0 & 73.6 \\
\oursrow \textbf{\method{} (Ours)} & \textbf{97.7} & \textbf{99.0} & \textbf{96.6} & \textbf{96.7} & \textbf{91.6} & \textbf{97.7} & \textbf{94.3} & \textbf{93.3} & \textbf{95.9} \\
\bottomrule
\end{tabular}

%% file: tables/gemma.tex
\centering\scriptsize\setlength{\tabcolsep}{1.9pt}
\caption{\textbf{\method{} has the best average on Gemma-4-31B.} Medical classification on BreakHis (BHis)~\cite{spanhol2016breakhis}, TBX11K (TBX)~\cite{liu2020tbx11k}, DDR~\cite{li2019ddr} and LAG~\cite{li2019lag}; medical VQA on PathVQA (Path)~\cite{he2021pathvqa}, SLAKE~\cite{liu2021slake}, VQA-RAD (RAD)~\cite{lau2018vqarad} and VQA-Med 2019 (Med)~\cite{benabacha2019vqamed}, answers scored by an LLM judge~\cite{li2026grading}, GPT-6 Luna; scored by rule, \method{} still leads RegMean (Supp.~A). Zero-shot and experts: accuracy (\%); merges: normalized (\%). Bold: best merge; underlined: second best.}
\label{tab:gemma}
\begin{tabular}{l|rrrrrrrr|r}
\toprule
Method & BHis & TBX & DDR & LAG & Path & SLAKE & RAD & Med & Avg \\
\midrule
Zero-shot & 61.4 & 63.1 & 57.1 & 76.5 & 39.1 & 66.6 & 63.9 & 40.6 & 58.5 \\
Fine-tuned & 88.3 & 98.2 & 71.3 & 93.7 & 43.6 & 72.2 & 60.8 & 44.4 & 71.6 \\
\midrule
Averaging & 75.9 & 67.0 & 85.1 & 89.1 & 94.0 & 92.8 & \underline{102.2} & 92.8 & 87.4 \\
TA & 87.3 & 81.5 & 92.7 & 79.9 & 90.2 & 93.5 & 97.4 & 93.7 & 89.5 \\
TIES & 92.0 & 77.2 & 91.4 & 83.5 & 93.8 & 91.8 & 98.2 & 92.3 & 90.0 \\
WTA & 75.1 & 74.6 & 88.9 & 77.1 & 92.9 & 93.9 & 101.8 & 93.7 & 87.3 \\
RegMean & \underline{98.5} & \underline{95.4} & \underline{96.1} & \textbf{97.3} & \underline{99.3} & \underline{94.4} & 101.5 & \textbf{95.9} & \underline{97.3} \\
AdaMerging & 86.4 & 74.1 & 89.8 & 91.7 & 93.1 & 93.5 & \textbf{105.1} & 89.6 & 90.4 \\
NSC & 86.4 & 75.4 & 91.7 & 86.7 & 92.6 & 93.1 & 101.8 & 91.0 & 89.8 \\
\oursrow \textbf{\method{} (Ours)} & \textbf{100.0} & \textbf{96.5} & \textbf{97.3} & \underline{96.9} & \textbf{99.4} & \textbf{98.7} & 101.5 & \underline{94.1} & \textbf{98.1} \\
\bottomrule
\end{tabular}

%% file: tables/llm.tex
\centering\scriptsize\setlength{\tabcolsep}{1.75pt}
\caption{\textbf{\method{} also has the best average on language-model adapters.} NLI: normalized accuracy (\%), the six KnOTS LoRAs~\cite{stoica2025knots} on Llama-3-8B~\cite{grattafiori2024llama3}. Gen.: normalized ROUGE-L (\%), Lots-of-LoRAs adapters~\cite{bruelgabrielsson2024compress} on Mistral-7B~\cite{jiang2023mistral}. Avg.: weight averaging; AdaM.: AdaMerging. Bold: best merge; underlined: second best.}
\label{tab:llm}
\begin{tabular}{l|r|rrrrrrrr}
\toprule
Pool & Base & Avg. & TA & TIES & WTA & AdaM. & NSC & RegMean & \method{} \\
\midrule
NLI, $K = 6$ & 51.0 & 80.3 & 90.2 & 89.1 & 86.3 & 77.7 & 86.7 & \underline{96.8} & \textbf{97.6} \\
Gen., $K = 8$ & 67.0 & 87.9 & 87.9 & 88.1 & 77.6 & 80.6 & 87.5 & \underline{97.4} & \textbf{98.3} \\
Gen., $K = 16$ & 65.8 & 79.7 & 79.7 & 80.0 & 72.7 & 75.1 & 78.5 & \underline{95.8} & \textbf{96.6} \\
Gen., $K = 32$ & 62.6 & 79.1 & 79.1 & 78.5 & 66.7 & 60.6 & 59.0 & \underline{90.8} & \textbf{93.0} \\
\bottomrule
\end{tabular}

%% file: tables/sites_ablate.tex
\begin{table*}[t]
\begin{minipage}[t]{0.36\textwidth}
\centering\small\setlength{\tabcolsep}{2.5pt}
\caption{\textbf{One merged adapter beats every site and the ten-site ensemble.} Qwen3-VL-2B, one dermoscopy VQA task over ten sites; accuracy (\%) on 1,306 test questions. On the held-out split~2, \method{} leads the best site by [+6.4, +9.7] points (95\% paired bootstrap interval).}
\label{tab:sites}
{\footnotesize\begin{tabular}{lrr}
\toprule
Method & Split 1 & Split 2 \\
\midrule
\multicolumn{3}{c}{\textit{Not merges}} \\
\midrule
Pooled LoRA (shared data) & 98.6 & 98.5 \\
Best single site & 90.0 & 88.7 \\
Other sites & 79.1--89.2 & 78.6--87.9 \\
Ten-site ensemble (10$\times$ cost) & 89.4 & 88.7 \\
\midrule
\multicolumn{3}{c}{\textit{Merged into one LoRA}} \\
\midrule
Weight averaging & 90.0 & 88.7 \\
TA~\cite{ilharco2023editing} & 94.6 & 93.5 \\
WTA~\cite{mou2026mergemedbench} & 92.6 & 91.7 \\
TIES~\cite{yadav2023ties} & 93.3 & 94.0 \\
AdaMerging~\cite{yang2024adamerging} & 87.6 & 92.5 \\
NSC~\cite{lee2026nsc} & 92.6 & 92.5 \\
RegMean~\cite{jin2023regmean} & 94.6 & 93.0 \\
\oursrow \textbf{\method{} (Ours)} & \textbf{96.6} & \textbf{96.8} \\
\bottomrule
\end{tabular}}
\end{minipage}\hfill
\begin{minipage}[t]{0.615\textwidth}
\centering\small\setlength{\tabcolsep}{3pt}
\caption{\textbf{Ablation: the label-free pick finds the best scale on every pool.} Normalized accuracy (sites: raw). Merger's Grams: base-model Grams on the merger's pool inputs. $\rho$: agreement against accuracy over the merges of \cref{fig:agree}. IVL: InternVL2-1B. Bold: best per column. $^\dagger$Best of $1/K$, 0.15 and 0.2.}
\label{tab:ablate}
{\footnotesize\begin{tabular}{lrrrrrrr}
\toprule
Variant & B/32 & L/14 & Qwen & IVL & Gemma & Sites 1 & Sites 2 \\
\midrule
\multicolumn{8}{c}{\textit{Grams $\times$ scale: accuracy (\%)}} \\
\midrule
RegMean (experts' Grams, $1/K$) & 71.0 & 78.2 & 97.6 & 91.1 & 97.3 & 94.6 & 93.0 \\
Base Grams, $1/K$ & 76.4 & 78.1 & 97.5 & 91.8 & 97.3 & 95.3 & 93.6 \\
Base Grams, $c = 0.2$ & 54.1 & 61.5 & \textbf{98.6} & \textbf{95.9} & 97.5 & 96.5 & 96.8 \\
\oursrow \textbf{\method{}, merger's Grams} & \textbf{80.2} & 81.6 & 98.4 & 95.6 & \textbf{98.5} & \textbf{97.1} & \textbf{97.1} \\
\oursrow \textbf{\method{} (base Grams, $c^\star$)} & 79.8 & \textbf{81.9} & \textbf{98.6} & \textbf{95.9} & 98.1 & 96.6 & 96.8 \\
\midrule
\multicolumn{8}{c}{\textit{Does agreement find the scale?}} \\
\midrule
Label-free pick $c^\star$ & 0.1 & 0.075 & 0.2 & 0.2 & 0.15 & 0.15 & 0.2 \\
Labeled pick & 0.1 & 0.075 & 0.2 & 0.2 & 0.15 & = $c^\star$ & = $c^\star$ \\
Best $c$ on test & 0.1 & 0.075 & 0.2 & 0.2 & 0.15$^\dagger$ & = $c^\star$ & 0.2 \\
$1/K$ & 0.125 & 0.125 & 0.125 & 0.125 & 0.125 & 0.1 & 0.1 \\
Spearman $\rho$ & 1.00 & 1.00 & 0.99 & 1.00 & 0.98 & 0.98 & 0.97 \\
\bottomrule
\end{tabular}}
\end{minipage}
\end{table*}

%% file: tables/cost.tex
\begin{table*}[t]
\centering\scriptsize\setlength{\tabcolsep}{3.2pt}
\caption{\textbf{\method{} needs no labels or gradients and runs in minutes.} Rounds: exchanges with the holders during the merge; Labels: labeled validation data (CLIP only). Message: the adapter, plus the Grams for RegMean and \method{}, sent once in fp32. Time: from the adapters to one merged adapter, Grams and pick included (VLMs: the pick on 50 pool questions per task). Qwen, IVL: Qwen3-VL-2B, InternVL2-1B.}
\label{tab:cost}
\begin{tabular}{lrrrrrrrrrrrrr}
\toprule
 & \multicolumn{3}{c}{Requirements} &  & \multicolumn{4}{c}{Message per holder (MB)} & \multicolumn{4}{c}{Time on one GPU (s)} &  \\
\cmidrule(lr){2-4} \cmidrule(lr){6-9} \cmidrule(lr){10-13}
Method & Grams & Labels & Grad. & Rounds & B/32 & L/14 & Qwen & IVL & B/32 & L/14 & Qwen & IVL & GPU mem. (GB) \\
\midrule
\multicolumn{14}{c}{\textit{Gradient-free}} \\
\midrule
TA, TIES, TSV-M, KnOTS-TIES & \xmark & \cmark & \xmark & 0 & 4.7 & 13 & 26 & 8.7 & $\le$1.4 & $\le$4.8 & $\le$55 & $\le$7.8 & 6.5 / 2.4 \\
RegMean~\cite{jin2023regmean} & \cmark & \xmark & \xmark & 1 & 61 & 214 & 965 & 163 & 0.8 & 3.3 & 50 & 42 & 11.2 / 6.1 \\
\oursrow \textbf{\method{} (Ours)}\tsub{a} & \cmark & \xmark & \xmark & 1 & 61 & 214 & 965 & 163 & 3.8 & 30 & 293 & 371 & 11.2 / 6.1 \\
\midrule
\multicolumn{14}{c}{\textit{Gradient-based}} \\
\midrule
AdaMerging~\cite{yang2024adamerging} & \xmark & \xmark & \cmark & 0 & 4.7 & 13 & 26 & 8.7 & 123 & 502 & 553 & 734 & 18.7 / 19.5 \\
NSC~\cite{lee2026nsc} & \xmark & \xmark\tsub{b} & \cmark & 0 & 4.7 & 13 & 26 & 8.7 & 27 & 184 & 523 & 699 & 18.8 / 19.5 \\
\bottomrule
\end{tabular}
\par\vspace{2pt}{\footnotesize\raggedright \tsub{a}Adapter only, and no round, when the merger computes the Grams; two rounds when the holders score the pick (\cref{sec:pick}). \tsub{b}On CLIP its initial coefficient is picked on labeled data.\par}
\end{table*}

%% file: tables/scorers.tex
\begin{table}[t]
\centering\footnotesize\setlength{\tabcolsep}{3.3pt}
\caption{\textbf{Only agreement finds the best scale on every pool.} Points of normalized test accuracy lost against the most accurate of the twelve scales of \cref{fig:scale}, with each scorer's pick in parentheses (Eq.~\ref{eq:pick} for agreement); same candidates and pool. Consistency: AdaMMS's rule~\cite{du2025adamms}, disagreement with the two neighboring scales. Entropy: of the merge's answers, which AdaMerging minimizes. KL: from each adapter's answers to the merge's. Feature distance: of the merge's head input from each adapter's, relative to the adapter's own change. All but agreement pick their lowest value.}
\label{tab:scorers}
\begin{tabular}{lrrrr}
\toprule
Scorer & B/32 & L/14 & Qwen & IVL \\
\midrule
\textbf{Agreement, \method{}} & \textbf{0.0 (.1)} & \textbf{0.0 (.075)} & \textbf{0.0 (.2)} & \textbf{0.0 (.2)} \\
Consistency, AdaMMS & 2.4 (.075) & 1.2 (.05) & 0.1 (.25) & 1.4 (.15) \\
Entropy & 13.2 (.025) & 5.4 (.025) & 0.1 (.25) & 9.7 (.3) \\
KL from each adapter & 25.7 (.2) & 7.9 (.15) & 0.0 (.2) & 0.0 (.2) \\
Feature distance & 8.5 (.15) & 3.7 (1/8) & 1.1 (1/8) & 1.4 (.15) \\
\bottomrule
\end{tabular}
\end{table}

%% file: sec/5_conclusion.tex
\section{Conclusion}
\label{sec:conclusion}

RegMean merges LoRA adapters in closed form without their training data, but fixes the scale of the merged update. We showed that the best scale lies on either side of RegMean's $1/K$ and moves with the adapters being merged, and that the merged model's agreement with each adapter on unlabeled inputs finds it. \method{} regresses once on statistics of the frozen base model, picks the scale by agreement and returns one exact LoRA. The pick also works without base-model Grams: it improves RegMean on the experts' own Grams on every pool, and on CLIP it finds LOT's labeled scale. \method{} has a higher average than every baseline we compare on every vision and vision-language model we test, from CLIP to Gemma-4-31B, and on four pools of language-model adapters. From ten sites with skewed shares of one task it builds an adapter that beats every site.

\paragraph{Limitations.} Like RegMean's, our Grams carry no formal privacy guarantee. Like AdaMerging, NSC and Expert Merging, the pick needs unlabeled inputs at the merger (or, for \method{}, one scoring round with the holders), and our runs draw them from each task's own distribution.

%% file: sec/X_suppl.tex
\clearpage
\maketitlesupplementary
\appendix
\makeatletter\let\Hy@backout\@gobble\makeatother

This supplement gives the details that the main paper points to: the setup of each benchmark (\cref{supp:setup}), our readings of the baselines and the spectral and LOT merges (\cref{supp:baselines}), the scale and its pick (\cref{supp:curves}: the proof of Observation~\ref{obs:bracket}, the numbers behind Fig.~\ref{fig:scale}, how agreement bounds accuracy, RegMean's own Grams, other label-free scorers, a scale computed from the message alone, and shrinkage), paired standard errors (\cref{supp:paired}), a natural-image pool with a spatial expert (\cref{supp:spatial}), a pool that adds the spatial expert to the medical ones (\cref{supp:medspatial}), how many unlabeled inputs \method{} needs (\cref{supp:sensitivity}) and qualitative examples, one test item per task (\cref{supp:gallery}).

\section{Setup}
\label{supp:setup}

\input{tables/supp_setup}

\Cref{tab:setup} summarizes the pools. Every holder records its Grams on its calibration inputs, one input at a time so that no padding enters, with the adapted layers in fp32. Every token of that forward pass enters: the image tokens on CLIP, and on the VLMs every image and prompt token, with the chat template's generation prompt; no label, answer or generated token does. The merger solves with a Cholesky factorization of $M + \epsilon I$. Projections that share an input share one Gram: query, key and value share one, the output projection has its own, and on Gemma-4-31B the gate and up projections share one and the down projection has its own. As for RegMean, a Gram is $d_{\text{in}} \times d_{\text{in}}$ per distinct input: in fp32, 0.15\,GB per holder on InternVL2-1B and 0.94\,GB on Qwen3-VL-2B. A base-model Gram serves every later merge that keeps the base model, the adapted modules and the input format, such as the prompt template; the scale is picked again for each merge.

\paragraph{CLIP.} The eight LoRA experts released with KnOTS~\cite{stoica2025knots} adapt the query, key, value and output projections of every vision block of CLIP ViT-B/32 and ViT-L/14 at rank 16. Scoring follows KnOTS: the normalized visual projection of the pooled vision output is matched against zero-shot text heads built from KnOTS's class names and templates, under fp16 autocast. KnOTS holds out 20\% of each test set as validation. We use its index for MNIST, SVHN and SUN397, whose public copies keep the order, and a seeded 20\% for the other five; the rest, 69,947 images, is our test set. From the held-out part we draw per task 64 unlabeled calibration images for the Grams, 150 unlabeled pool images for the label-free pick, AdaMerging and NSC, and 150 labeled validation images for tuning the baselines. On KnOTS's test split every ViT-B/32 expert is within 0.63 points of KnOTS's reported accuracy and every ViT-L/14 expert within 1.1 points.

\paragraph{Medical VLMs.} The MergeMedBench experts~\cite{mou2026mergemedbench} are LoRAs with $r = 16$ and $\alpha = 32$ on the query, key, value and output projections of the language model only. The InternVL2-1B experts share one initialization and their updates overlap (mean pairwise cosine 0.42); the Qwen3-VL-2B experts start from separate random initializations and are nearly orthogonal (0.03). Each modality has 300 test questions, 50 labeled validation questions, 250 unlabeled pool questions and 64 calibration questions, drawn at random with fixed seeds from MergeMedBench's released test set, and no test image appears in another split. MergeMedBench released no experts for Qwen3-VL-4B, so we trained one per modality with the released experts' LoRA configuration on 2,000 OmniMedVQA training questions whose images are in no test split; its label-free pick uses 50 of the 250 pool questions per modality. Answers are parsed from the format \texttt{<answer> X </answer>}, and decoding is greedy. Our InternVL2-1B evaluation code matches the official code: identical token ids and logits within 0.003.

\paragraph{Agreement.} Two answers agree when they are the same under the task's scoring rule. CLIP: the same predicted class. Medical VLMs: the same option letter parsed from \texttt{<answer> X </answer>}; an off-format answer agrees only with another off-format answer. OptMerge pool (Supp.~E): each benchmark's own extraction, that is normalized text for VizWiz, TextVQA, GQA and OCR-VQA, numbers within 5\% for ChartQA, and boxes with IoU of at least 0.5 for RefCOCO; responses with nothing to extract agree only if their normalized texts match. Gemma-4-31B: exact match after lowercasing, collapsing whitespace and stripping end punctuation, reading the JSON label or answer field when present; a missing answer is a disagreement. NLI: the same predicted label. Generation: ROUGE-L of the merge's output against the adapter's, so agreement is graded rather than all or nothing.

\paragraph{Gemma-4-31B.} The eight adapters, trained with the recipe of the fine-tuned comparators of~\cite{zhao2026rgicl}, are LoRAs with $r = 16$ and $\alpha = 32$ on Gemma-4-31B-it~\cite{gemma2026gemma4}. The classification adapters (BreakHis, TBX11K, DDR, LAG) adapt the decoder's attention and MLP, and the VQA adapters (PathVQA, SLAKE, VQA-RAD, VQA-Med) also the vision projector; every adapted matrix is merged by the same rule, with one shared $c$. At the vision projector, which only the four VQA adapters adapt, the classification adapters count as zero updates: their Grams still enter $M$, so on their inputs the regression keeps the base projector. Calibration uses 64 training inputs per task, the pick uses 100 validation questions per task, and test uses the full test sets, 16,515 questions in all. The pick reads no validation label; the labels enter only the validation curve of Fig.~\ref{fig:scale} and the labeled pick of \cref{tab:ablate}. The judged test answers cover three scales, $1/K$, the pick and 0.2. GPT-6 Luna (OpenAI API model \texttt{gpt-6-luna}; reasoning effort none, temperature 0, seed 3407 plus the row index, JSON output, at most 512 tokens; \cite{zhao2026rgicl} used \texttt{gpt-5.4-mini-2026-03-17}) judges every VQA answer against its reference as an LLM judge~\cite{li2026grading}, with the prompt of~\cite{zhao2026rgicl} (\cref{tab:judge}); an answer is correct when the judge gives it full semantic accuracy. Scoring the same answers by rule instead (exact match, and token recall for open answers outside VQA-Med) gives \method{} 98.3 and RegMean 97.2 on the average of \cref{tab:gemma}.

\input{tables/supp_judge_prompt}

\paragraph{Language models.} \emph{NLI:} the six LoRAs released with KnOTS~\cite{stoica2025knots}, $r = 16$ and $\alpha = 16$ on the query, key, value and output projections of Llama-3-8B~\cite{grattafiori2024llama3}, each with its task's trained three-way classification head, for SNLI~\cite{bowman2015snli}, MNLI~\cite{williams2018mnli}, SICK~\cite{marelli2014sick}, QNLI~\cite{rajpurkar2016squad,wang2019glue}, RTE~\cite{dagan2005rte,wang2019glue} and SciTail~\cite{khot2018scitail}. The merged model answers each task through that task's head. Test is KnOTS's test split, 29,216 questions; 64 calibration and 250 pool inputs per task and 1,139 labeled validation questions come from outside it. Every expert is within 0.14 points of KnOTS's reported accuracy. \emph{Generation:} adapters released with Lots-of-LoRAs~\cite{bruelgabrielsson2024compress}, $r = 16$ and $\alpha = 32$ on the query, key and value projections of Mistral-7B-Instruct-v0.2~\cite{jiang2023mistral}, for Super-NaturalInstructions tasks~\cite{wang2022superni}, all prompted in the dataset's own format. The eight tasks at $K = 8$ are XLSum and Amazon review titles, WebNLG and NYC data-to-text, SQuAD question generation~\cite{rajpurkar2016squad}, ParaNMT paraphrases, CommonGen and story endings; test is the first 300 rows of each released test split, and 64 calibration and 50 pool prompts from the training split and 50 labeled validation prompts come from outside it. At $K = 16$ and 32 we add 8 and then 16 more adapters of the same release: Super-NaturalInstructions tasks 127, 111, 574, 68, 955, 184, 275 and 1401, then 1660, 82, 670, 132, 565, 1603, 413, 381, 67, 269, 1714, 223, 1407, 1340, 1669 and 568. They are the first, in a random order of the release's generation tasks, whose expert beats the base model by at least 5 ROUGE-L points and 2 paired standard errors on the task's 100 test rows. Every task then has 100 test, 25 pool and 64 calibration prompts. Decoding is greedy, and the score is ROUGE-L F1 on the text before the first newline, best over a row's references, divided by the expert's on the same rows.

\paragraph{Ten sites.} The questions are written exactly as MergeMedBench writes them and split by image, so that no image is in two sites, and no site holds an image of MergeMedBench's test file. The seven image sources are Fitzpatrick 17k~\cite{groh2021fitzpatrick}, ISIC 2016~\cite{gutman2016isic}, ISIC 2018~\cite{codella2019isic,tschandl2018ham10000}, ISIC 2019~\cite{combalia2019bcn20000,tschandl2018ham10000}, ISIC 2020~\cite{rotemberg2021isic2020}, a monkeypox skin image database~\cite{islam2022monkeypox} and PAD-UFES-20~\cite{pacheco2020padufes}. All seven are public research releases; the monkeypox images were gathered from public web pages by that database's authors. We use every source for research only, under its terms, and redistribute no image. Each image source is spread over the ten sites with Dirichlet(0.1) proportions, each image going to a site still below 20\% of all questions. Split~1 takes the first seed whose smallest site has at least 100 training questions; its sites hold 110 to 733 training questions (3,501 in all), most of them from one or two of the seven sources. Split~2 is a new Dirichlet(0.1) draw, again with no site above 20\%, with 3,503 training questions. Each site keeps 25\% of its questions as its unlabeled pool and 32 as calibration inputs. Each site trains a LoRA with $r = 16$, $\alpha = 32$ and dropout 0.1 on the query, key, value and output projections, with AdamW at $2 \times 10^{-4}$, 3\% warmup then linear decay, two epochs and eight questions per step, and its own seed. The pooled LoRA uses the same recipe on all sites' training data. Every model is scored on MergeMedBench's 1,306 dermoscopy test questions. The site training counts exclude each site's pool and calibration questions, and the ten-site ensemble answers each question with the option of highest probability averaged over the ten site models. Equal entries in \cref{tab:sites} are equal totals of different answers: on split~2 the best site, weight averaging and the ensemble each answer 1,159 of the 1,306 test questions correctly, but score 98.5, 87.8 and 86.3 on the ISIC 2016 images.

\paragraph{Scales and seeds.} Every pool of \cref{tab:curves} scores its twelve scales, and CLIP also 0.75 and 1; on the VLMs every scale is scored on the pool in fp32, and bf16 scoring at $1/8$ and 0.2 to 0.5 picks the same scale. The sites score $c \in \{0.05, 0.1, 0.15, 0.2, 0.3, 0.4\}$ on split~1 and the four scales up to 0.2 on split~2, for the regression and for the plain sum of the adapters, each with equal or size weights: 24 and 16 candidates; the pick chose the regression with size weights at 0.15 on split~1 and with equal weights at 0.2 on split~2. Adding 0.3 and 0.4 on split~2 leaves the pick unchanged, and each of the eight new candidates scores below it on test. CLIP's validation part for Cars, DTD, EuroSAT, GTSRB and RESISC45 is drawn with NumPy seed (1000, task index) and divided into calibration, pool and validation images with seed (0, task index). The site splits use Dirichlet seeds 93 and 187, the first seeds from 0 and from 94 whose smallest site trains on at least 100 questions. The $K$ sweep of Fig.~\ref{fig:ksweep} scores $1/K$ and the grid $\{0.1, 0.15, 0.2, 0.25, 0.3, 0.4, 0.5, 0.65, 0.8, 1\}$, from the largest grid value below $1/K$ upward, and stops once pool agreement falls 2 points below its best, so the pick can exceed 0.5.

\paragraph{Cost.} The messages of \cref{tab:cost} count the adapter's factors and every distinct Gram in full, in fp32. A holder that feeds the secure sum of \cref{sec:message} sends its Gram masked, at the same size. A base-model Gram is recorded once, so later merges with it cost only the solve and the pick. The times of \cref{tab:cost} start from loaded models and include recording the Grams, the solve and, for \method{}, the pick. On the VLMs the pick scores 50 pool questions per task, which pick the same scale as 250 (\cref{supp:sensitivity}). The times exclude the vision tower's pass over the calibration images, which RegMean shares, and any transfer between holders and merger.

\section{Baselines}
\label{supp:baselines}

Where a paper leaves a detail open we state the reading we used. On CLIP the hyperparameters of task arithmetic, TIES, TSV-M and KnOTS-TIES and NSC's initial coefficient are picked on the labeled validation images, as KnOTS does; RegMean runs at $c = 1/K$ with shrinkage 0.9, and AdaMerging at its published settings.

\paragraph{RegMean~\cite{jin2023regmean}.} Grams recorded inside each expert's own fine-tuned model on the same calibration inputs as \method{}'s, as per-token means like \method{}'s, so every holder weighs equally in $M$ whatever its token count, with the same shrinkage and ridge, at $c = 1/K$; Tab.~\ref{tab:experts} gives its gain at other scales. The shrinkage, 0.9, is the same for RegMean and \method{} and tuned for neither, so \cref{tab:ablate} compares Gram sources and scales at one shrinkage.

\paragraph{Weights-only merges.} Task arithmetic~\cite{ilharco2023editing}: $c$ on a grid, picked on validation on CLIP and by Eq.~(\ref{eq:pick}) on the medical models; on the sites the pick also chooses equal or size weights, as \method{}'s does. TIES~\cite{yadav2023ties}: density and $\lambda$ picked on validation on CLIP; density 0.2 and $\lambda = 1$, MergeMedBench's setting, on the medical VLMs and the sites; density 0.2 and $\lambda \in \{0.5, 1, 1.5\}$ picked by Eq.~(\ref{eq:pick}) on Gemma-4-31B. TSV-M~\cite{gargiulo2025tsv}: each update keeps its 16 singular triplets, since TSV's $k = d/K$ would exceed the rank. KnOTS-TIES~\cite{stoica2025knots}: KnOTS's global magnitude mask over all modules. Winner-take-all~\cite{mou2026mergemedbench}: the MergeMedBench method on the LoRA factors; our InternVL2-1B result is within 0.3 points of the published one. Our task arithmetic, TIES and KnOTS-TIES on CLIP reproduce KnOTS's published numbers within 0.4 points. On the language models, task arithmetic's $c$ is picked by Eq.~(\ref{eq:pick}), and TIES's density and $\lambda$ on labeled validation data, from $\{0.2, 0.4, 0.6\} \times \{0.5, 1, 1.5\}$ on NLI and from $\{0.2, 0.6\} \times \{0.5, 1\}$ on generation; at $K = 16$ and 32, which have no labeled validation split, TIES's setting is picked by Eq.~(\ref{eq:pick}).

\paragraph{AdaMerging~\cite{yang2024adamerging}.} Layer-wise: one coefficient per expert per block, shared by the block's adapted projections, initialized at 0.3, clamped to $[0, 1]$ and trained with Adam at learning rate $10^{-3}$ for 500 steps to minimize prediction entropy on the unlabeled pool. On CLIP the entropy is that of the zero-shot head's softmax, with 16 pool images per task per step; on the VLMs, that of the softmax over the option letters at the answer position, with one pool question per modality per step. Gemma-4-31B's tasks answer in free text, so there the entropy is that of the merged model's next-token distribution over its own greedy answer (at most 32 tokens), averaged over the answer's positions, with one pool question per task per step; the vision projector, adapted by the VQA adapters, has its own coefficient. On the NLI adapters the entropy is that of each task head's softmax, with 16 pool inputs per task per step. On the generation adapters it is the next-token entropy along the merged model's own greedy answer, as on Gemma-4-31B, with one pool prompt per task per step.

\paragraph{NSC~\cite{lee2026nsc}.} One unconstrained coefficient per expert per adapted module, trained to minimize, at the target modules, the share of each target's input that lies outside the row space of its task's down-projection $A_k$, in the paper's form, which does not depend on the LoRA scale $s$ (the released code folds $s$ into $A_k$ but not into the inverse of $A_kA_k^\top$, which multiplies the in-span share by $s^2$, 4 on the VLMs). On CLIP we follow the paper's ViT setting: the output projections of the last quarter of the blocks, AdamW at $10^{-3}$, 100 steps and 32 pool images per task per step; the initial coefficient, 0.4 in the paper, is picked on validation from $\{1/8, 0.2, 0.3\}$ and is 1/8 on both backbones. The released code covers the paper's LLaVA setting, and on the VLMs we use its optimizer settings: AdamW at $3 \times 10^{-4}$, 500 steps, one pool question per modality per step and initial coefficient $2.4/K = 0.3$. We target the output projections there as well, the best target in the paper's ablation; the code targets the value projection, since its LLaVA adapters cover only the query and value projections. Gemma-4-31B uses the VLM settings, with the output projections of the last quarter of the decoder layers as targets. The language models use them too: the targets are the output projections of the last quarter of the layers on the NLI adapters and the value projections on Mistral-7B, whose adapters do not adapt the output projection; on NLI, 16 pool inputs per task per step. Over the 500 steps the null-space share barely moves (0.995 to 0.992 at $K = 32$) while the mean coefficient falls by 0.06 to 0.08 on generation at every $K$; at $K = 32$, which starts from $2.4/32 = 0.075$, more than half end negative, and the merge scores below the base model.

Both learned methods read the same unlabeled pool as \method{}'s pick and return an exact LoRA; on the VLMs they train in fp32, on Gemma-4-31B with bf16 weights and fp32 coefficients and losses. Both start from a plain weighted sum, AdaMerging from task arithmetic at 0.3 and NSC from weight averaging on CLIP and from task arithmetic at 0.3 on the VLMs, and the tables report their trained coefficients. AdaMerging improves on its start on both CLIP backbones; on InternVL2-1B it halves its entropy objective while its accuracy falls, and NSC's objective changes by less than 0.1\%. Their start on the VLMs, task arithmetic at 0.3, scores 93.3 on Qwen3-VL-2B and 81.7 on InternVL2-1B, 5.3 and 14.2 points below \method{}. On InternVL2-1B, 15.8\% of AdaMerging's answers and 9.1\% of NSC's fall outside the required format.

\paragraph{Expert Merging~\cite{zhang2025expertmerging}.} One coefficient per expert per adapted weight matrix, kept in $(0, 1)$ by a sigmoid and initialized at the paper's prior, trained so that on task $k$'s unlabeled pool questions the merge matches expert $k$: the soft-target cross-entropy of their output distributions at every position, the mean squared distance of their middle-layer hidden states, and an $\ell_1$ pull toward the prior. We use the paper's settings for InternVL2.5 on InternVL2-1B (5 questions per task, prior 0.1, pull weight 5) and for Qwen2-VL on both Qwen3-VL models (10, 0.3 and 0.8), the released code's defaults otherwise (AdamW at 0.5, gradient accumulation 8, one epoch, temperature 1), and fp32. \cref{tab:medvlm} reports it trained on the 250 pool questions per task that the other learned merges read: 5.6, 3.6 and 22.3 points below \method{} on Qwen3-VL-2B, Qwen3-VL-4B and InternVL2-1B (paired SEs 0.48, 0.41 and 0.87). With the paper's question counts it scores 92.7, 94.7 and 67.8.

\input{tables/supp_spectral}

\paragraph{Spectral and LOT merges.} We also ran LOT Merging~\cite{sun2025lot}, Core Space~\cite{panariello2025core}, Iso-C and Iso-CTS~\cite{marczak2025isoc} on our CLIP ViT-B/32 adapters, splits and scoring (\cref{tab:spectral}). LOT Merging is the regression on the experts' own inputs without shrinkage, solved by pseudo-inverse and scaled by one global coefficient $\lambda$. As a baseline in \cref{tab:clip,tab:medvlm} it runs at $\lambda = 1$, which is $c = 1/K$, on the Grams that RegMean uses. On our labeled validation images, $\lambda = 1.2$, LOT's published value for ViT-B/32, scores 8.5 points below $\lambda = 1$. Agreement picks $\lambda = 0.8$ for LOT, the scale that labeled validation data pick. Iso-C, Iso-CTS and Core Space (TSV and Iso-C in core space, its released configuration) are ports of the released code, each scaled by a coefficient that the released linear search tunes on labeled validation data, stopped after two values below the best. On LoRA updates Iso-CTS's task-specific subspace holds only rounding noise, since the summed rank is below the size of its common subspace. With these merges among the candidates, the merge that agrees most with the adapters on the pool is still the most accurate.

\section{The scale and its pick}
\label{supp:curves}

\paragraph{Proof of Observation~\ref{obs:bracket}.} (i) Here $M = \Gt$, so $\Gt_k M^{-1} = I$; with the ridge the map is $\Gt(\Gt + \epsilon I)^{-1}$, whose eigenvalues lie in $[1 - \epsilon/\lambda_{\min}, 1]$, and shrinkage keeps $\lambda_{\min} \ge 0.1 \min_i (G)_{ii}$. (ii) $M$ is block diagonal over the $S_j$ and $M^{-1}$, read as the pseudo-inverse on $\oplus_j S_j$, acts on $S_k$ as $KG_k^{+}$; for $x \in S_k$ this gives $G_k M^{-1}x = Kx$ and $G_j M^{-1}x = 0$ for $j \neq k$. \hfill$\square$ With the ridge, $M + \epsilon I$ is positive definite whenever a layer receives any nonzero input, so (ii) is the raw-Gram limit of the implemented solve.

\input{tables/supp_curves}

\Cref{tab:curves} lists the points of Fig.~\ref{fig:scale}. On every pool \method{}'s pick is the most accurate scored scale and equals the pick on labeled validation data (Tab.~\ref{tab:ablate}). On CLIP the best scale lies below $1/K$. On the medical VLMs and Gemma-4-31B it lies above $1/K$, and accuracy drops at every smaller scale down to 0.025.

\paragraph{Agreement and accuracy.} Fix task $k$ with one correct answer per input. Let $a_k > 0$ be adapter $k$'s accuracy on $U_k$, and $\mathrm{acc}_k(c)$ and $g_k(c)$ the merge's accuracy on $U_k$ and its agreement with adapter $k$ there. Where the merge agrees with a correct adapter it is correct, and where it is correct but disagrees, the adapter is wrong, so $|\mathrm{acc}_k - g_k| \le 1 - a_k$. Write $N(c) = \frac{1}{K}\sum_k \mathrm{acc}_k(c)/a_k$ for normalized pool accuracy, $A(c)$ for the agreement of Eq.~(\ref{eq:pick}) and $m = \frac{1}{K}\sum_k (1 - a_k)/a_k$. Since $g_k \le 1$, $A - m \le N \le A + 2m$, so the agreement pick $c^\star$ satisfies $N(c) - N(c^\star) \le 3m$ for every $c$. With the experts' test accuracies plugged in for $a_k$, $3m$ is 1.2 points on InternVL2-1B and 1.5 on Qwen3-VL-2B and Qwen3-VL-4B. Measured directly, the pick's loss against the best scale on test is zero on every pool of Tab.~\ref{tab:ablate} (also Fig.~\ref{fig:agree}).

\input{tables/supp_experts}

\paragraph{The experts' Grams.} \Cref{tab:experts} rescales RegMean's merge on the experts' own Grams and gives the points gained over $1/K$. Agreement picks a better scale than $1/K$ for these Grams as well, on every pool, so the pick also improves RegMean itself, by 0.9 to 5.0 points, while a fixed $c = 0.2$ loses more than 20 points on CLIP, on Gemma-4-31B and on split~2. Base-model Grams depend on no adapter: one message serves every later merge, and a merger with unlabeled inputs can compute it alone.

\paragraph{Other label-free scorers.} \Cref{tab:scorers} in the main paper sets agreement against four other label-free scorers on the same twelve candidates and pools. Agreement picks the most accurate scale on all four pools; every other scorer misses it on both CLIP backbones, by 1.2 to 25.7 points. Consistency is AdaMMS's rule~\cite{du2025adamms}: each candidate's disagreement with the two scales next to it on the grid, lowest wins, and the two end scales serve only as neighbors. It also misses on both VLMs, by 0.1 and 1.4 points. Entropy has no fixed direction: on CLIP it falls as $c$ shrinks and picks the smallest scale, and on InternVL2-1B it falls as $c$ grows past the best scale and picks 0.3, 9.7 points below it. KL from each adapter picks a scale above the best on CLIP. Feature distance compares the merge's head input with each adapter's on the pool inputs, in fp32: the normalized image embedding on CLIP, and the final norm's output at the answer position on the VLMs. With $e$ the adapter's change from the base model and $m$ the merge's, it is $\lVert m - e \rVert / \lVert e \rVert$, summed over a task's inputs and averaged over tasks. It is smallest at 0.15, $1/8$, $1/8$ and 0.15 on the four pools, next to the layerwise optimum $1/K$, while accuracy peaks at 0.1, 0.075, 0.2 and 0.2. Without normalization it picks the same four scales, and relative to the base model's features the same except 0.15 on ViT-L/14.

\begin{figure}[t]
\centering
\includegraphics[width=\linewidth]{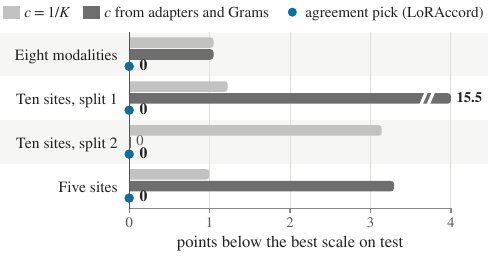}
\caption{\textbf{Agreement finds the best scale; the adapters and Grams alone do not.} Points lost against the best scale on test, Qwen3-VL-2B (five sites: the ten-site task split over five sites). From adapters and Grams: the $c$ that best fits each adapter's own update on its base-model calibration inputs, with the other adapters' terms weighted by one factor fit on 76 other merges.}
\label{fig:scalepred}
\end{figure}

\paragraph{A scale from the message alone.} The merger could also try to compute $c$ from what the holders send, without scoring candidates. For each layer, the scale that best fits each adapter's own update on its base-model calibration inputs, with the other adapters' terms weighted by one factor, can be computed from the adapters and Grams. We fit that factor and the aggregation over layers on the 76 merges of the $K$ sweeps. On four held-out Qwen3-VL-2B merges (\cref{fig:scalepred}) the computed scale loses 19.8 points in all against the best scale on test, and $1/K$ loses 6.4; the agreement pick loses none.

\paragraph{Shrinkage.} The off-diagonal shrinkage, 0.9 for RegMean and \method{} alike, leaves the pick unchanged. On ViT-B/32, with the same recorded Grams solved at shrinkage 0.5, 0.9 and none (the pseudo-inverse with LOT's cutoff), agreement picks 0.1 each time, the most accurate of the twelve scales. On Qwen3-VL-2B, with 50 pool questions per task and scales from 0.1 to 0.4, agreement picks 0.2 at all three. At every shrinkage, the best scale lies below $1/K$ on ViT-B/32 and above it on Qwen3-VL-2B.

\section{Paired differences}
\label{supp:paired}

\input{tables/supp_paired}

Differences between two merges on the same test items carry paired bootstrap standard errors, with 2,000 draws that resample test items within each task (multi-task) or image source (multi-site). A few test questions share an image, 2,400 questions on 2,383 images on the medical VLMs and 1,306 on 1,278 on the sites, so we also resampled whole images, every question of a drawn image entering together; the standard errors move by at most 0.02 points. \Cref{tab:paired} lists the differences behind the main text.

\section{Natural-image VQA with a spatial expert}
\label{supp:spatial}

\input{tables/supp_spatial}
\input{tables/supp_medspatial}

\method{} also merges general-domain experts on a 7B VLM. It gives the best average over eight of OptMerge's benchmarks without labels and keeps the spatial expert's accuracy on 3DSRBench (\cref{tab:spatial}).

\paragraph{Pool.} OptMerge~\cite{wei2026optmerge} released five Qwen2-VL-7B~\cite{wang2024qwen2vl} experts for general VQA, geometry, charts, OCR and grounding, each trained as a rank-8 LoRA on the language model and released merged into the base weights. We merge four of them and leave out the geometry expert, which scores below the zero-shot model on its own labeled validation data (26.0 against 40.0). We recover each LoRA as the rank-8 truncated SVD of its difference from the base model. A fifth expert covers 3D spatial reasoning. 3DSRBench~\cite{ma2025threedsrbench} has no training split, so we split its MS-COCO~\cite{lin2014coco} questions by image: 600 images (1,704 questions) are held out for test, and the questions on 1,107 other images (3,127) train a rank-8 LoRA with $\alpha = 16$ on the same 196 projections as OptMerge's experts. Training runs two epochs with AdamW at $10^{-4}$, a cosine schedule and batch 16, with the loss on the answer letter and the options rotated at random each time a question is drawn. Every task's calibration, pool and validation questions (64, 100 and 100) come from images outside its test set.

\paragraph{Scoring.} The eight benchmarks, VizWiz~\cite{gurari2018vizwiz}, GQA~\cite{hudson2019gqa}, ChartQA~\cite{masry2022chartqa}, TextVQA~\cite{singh2019textvqa}, OCR-VQA~\cite{mishra2019ocrvqa} and RefCOCO, RefCOCO+ and RefCOCOg~\cite{kazemzadeh2014referitgame,yu2016refcoco,mao2016refcocog} (2,350 test questions), are scored by rule-based answer extraction, where OptMerge used GPT-4o-mini; Avg is their mean. 3DSRBench uses its CircularEval: a question counts as correct only when every cyclic rotation of its options is answered correctly.

\paragraph{Merges.} \method{} uses base-model Grams and the label-free pick on the pool, which selects $c = 0.25$, above $1/K = 1/5$. RegMean is at $c = 1/K$. Task arithmetic and TIES (density 0.2) pick $\lambda$ on labeled validation data from OptMerge's grid $\{0.1, 0.3, 0.5, 0.7, 1, 1.5\}$.

\paragraph{Results.} Over the eight benchmarks, \method{} is 2.8 points above task arithmetic and 2.4 above TIES, both tuned with labels, and 0.8 above RegMean (paired SEs over test items 0.62, 0.62 and 0.37). On 3DSRBench it is the only merge that reaches the spatial expert's accuracy.

\section{Medical and spatial experts in one pool}
\label{supp:medspatial}

\input{tables/supp_sensitivity}

\method{} merges experts from different domains into one adapter. With a spatial-reasoning expert added to the eight medical experts of Qwen3-VL-2B, it gives the best average over the nine tasks and keeps the spatial expert's accuracy on 3DSRBench (\cref{tab:medspatial}).

\paragraph{Pool.} The ninth expert is trained on the 3DSRBench training split of \cref{supp:spatial} (3,127 questions on 1,107 images), with each question written in MergeMedBench's multiple-choice format. It follows the recipe of the released medical experts: a rank-16 LoRA with $\alpha = 32$ and dropout 0.1 on the query, key, value and output projections of the language model, two epochs of AdamW at $2 \times 10^{-4}$ with 3\% warmup and linear decay, and eight questions per step, with the options rotated at random in each epoch. The spatial task's 64 calibration and 100 pool questions come from images outside its test set.

\paragraph{Merges.} All merges are label-free, as on the medical VLMs in the main paper. \method{} uses base-model Grams of 64 calibration questions per task and picks its scale by agreement on the pool, 250 questions per modality and the 100 spatial ones, which selects $c = 0.15$, above $1/K = 1/9$. Medical tasks are scored on the test questions of \cref{tab:medvlm}, 3DSRBench by CircularEval. Every merge answers every question in the required format.

\paragraph{Results.} Over the nine tasks, \method{} is 0.8 points above RegMean, 4.5 above TIES and 5.4 above task arithmetic (paired SEs 0.28, 0.48 and 0.48). It also stays closest to each expert: on the spatial test prompts it gives the spatial expert's answer 85\% of the time, against 77\% for RegMean and 71\% for TIES.

\section{How many unlabeled inputs}
\label{supp:sensitivity}

\begin{figure*}[t]
\centering
\includegraphics[width=\textwidth]{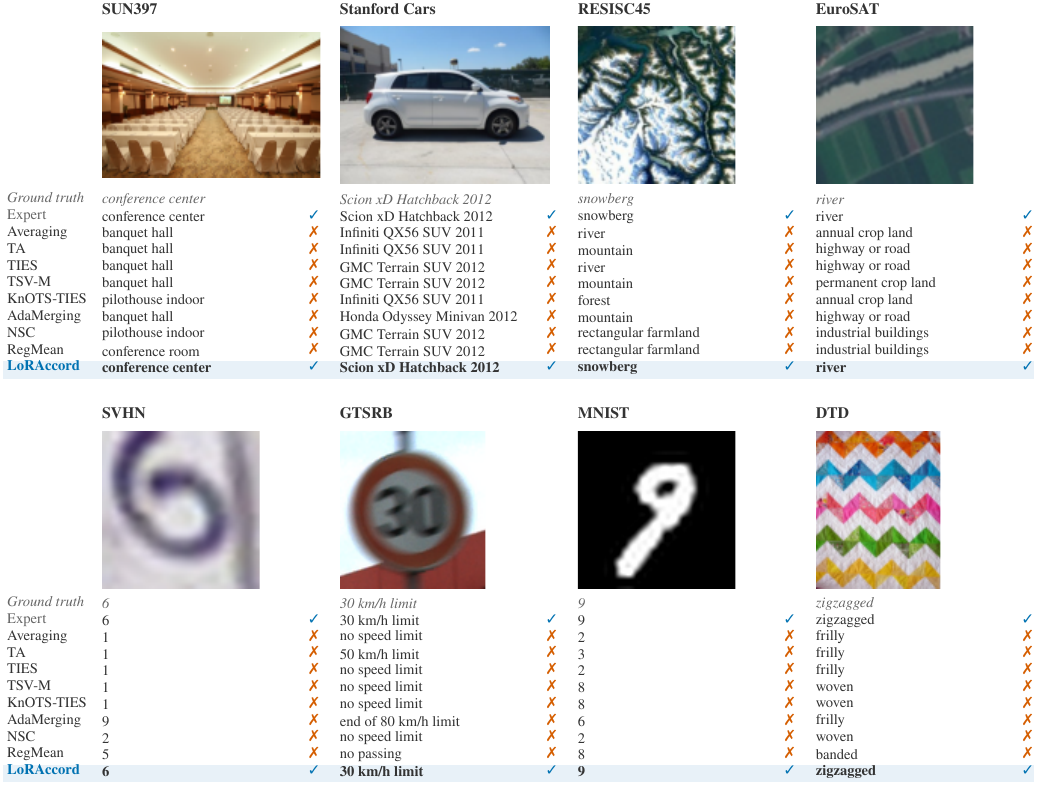}
\caption{\textbf{Answers on each CLIP task.} One test image per task of \cref{tab:clip} (ViT-B/32), with the class that the task's expert, each merge and \method{} predict. Long GTSRB and EuroSAT class names are shortened.}
\label{fig:gal_clip}
\end{figure*}

\method{} needs few unlabeled inputs, for the Grams and for the pick (\cref{tab:sensitivity}).

\paragraph{Calibration inputs.} On InternVL2-1B we recorded the base-model Grams on random subsets of $n$ calibration questions per task, three draws for each $n$, and merged at $c = 0.2$. Accuracy levels off at 32 questions per task, where the spread between draws also shrinks; the main experiments use 64.

\paragraph{Pick pool.} We re-applied the pick to random subsets of the ten sites' saved pool answers on split~1, where every candidate was also scored on test. With 102 pool questions in all, the pick scores above the best single site in 99.9\% of draws. On the medical VLMs, 50 pool questions per task pick the same scale as the full pool of 250.

\paragraph{Resampled pools.} We also redrew each pool from its own items: 2,000 bootstrap draws resample every task's (site's) pool items with replacement, the same draw for every candidate, and re-apply the pick to the saved answers. The pick's mean loss against the most accurate candidate on test is 0.21 points on ViT-B/32, 0.01 on ViT-L/14, 0.11 on Qwen3-VL-2B and 0.00 on InternVL2-1B, and 0.09 and 0.40 on the two site splits.

\section{Qualitative examples}
\label{supp:gallery}

\begin{figure*}[p]
\centering
\includegraphics[width=\textwidth]{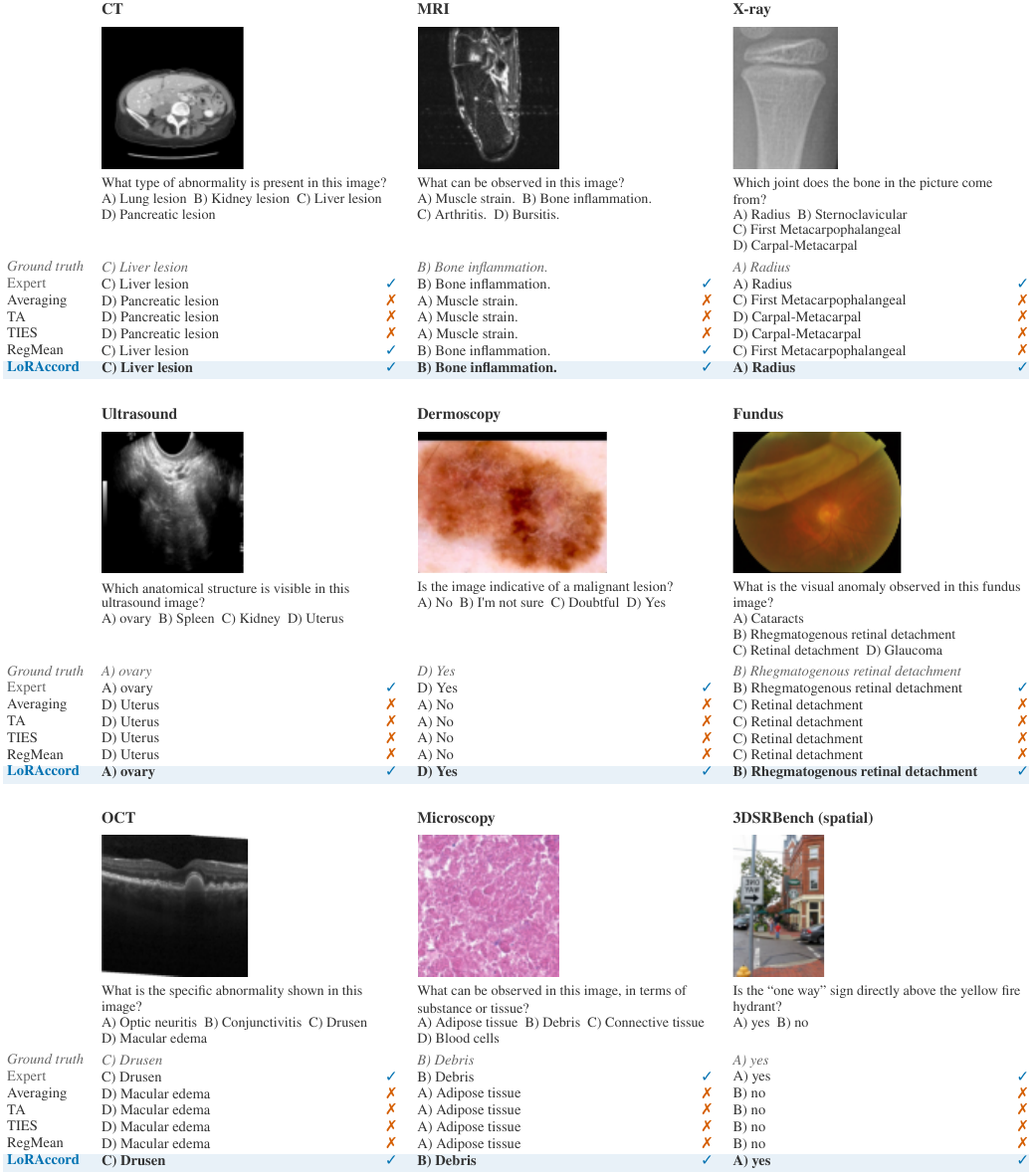}
\caption{\textbf{Answers on each medical and spatial task.} One test question per task of the nine-adapter pool of \cref{tab:medspatial} (Qwen3-VL-2B). 3DSRBench counts a question only when every rotation of the options is answered correctly; the letter shown answers the original order, and the merges marked wrong pick the wrong option under both orders.}
\label{fig:gal_med}
\end{figure*}

\begin{figure*}[p]
\centering
\includegraphics[width=\textwidth]{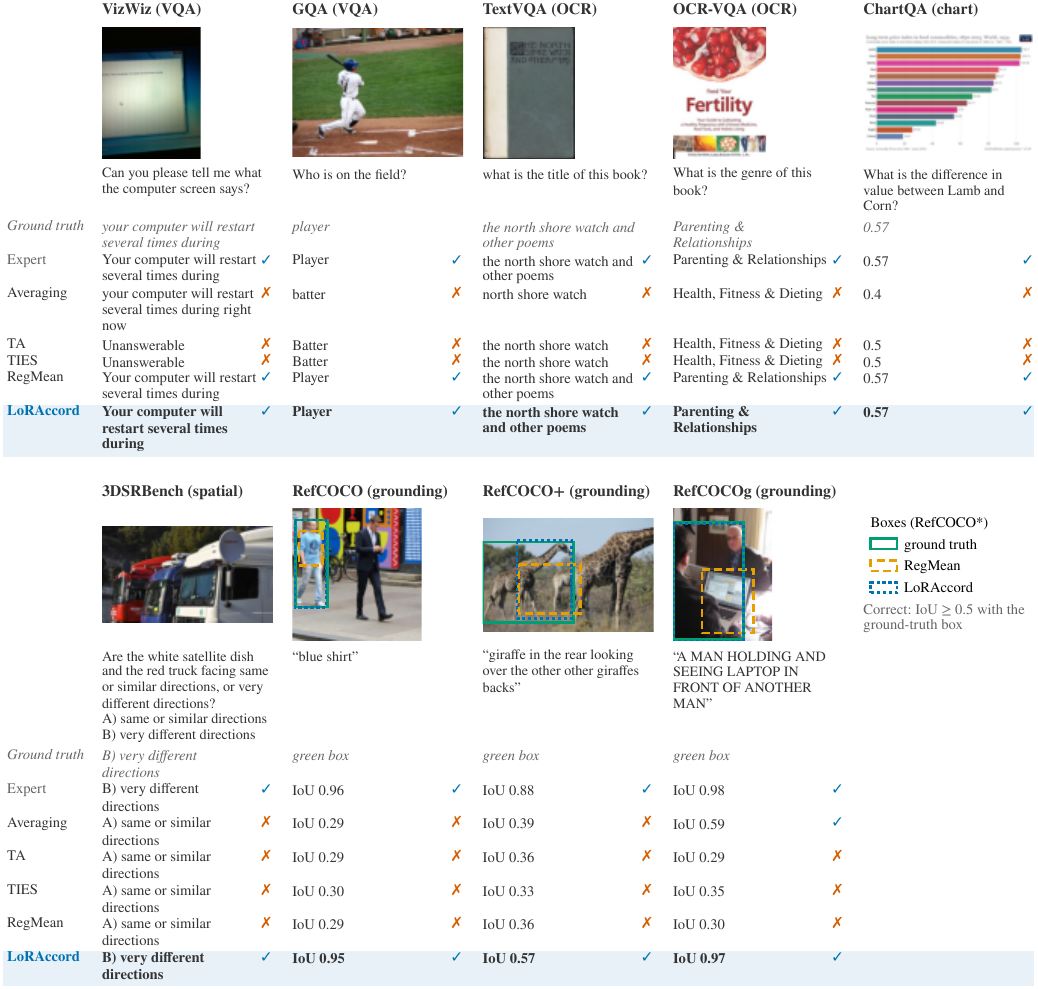}
\caption{\textbf{Answers on each natural-image benchmark.} One test question per benchmark of the pool of \cref{tab:spatial} (Qwen2-VL-7B), with the answers as the models wrote them; the 3DSRBench question is shortened. VizWiz and TextVQA show the most frequent of the ten reference answers. RefCOCO*: IoU of each box with the ground truth (green); the boxes of RegMean and \method{} are drawn. Correctness follows each benchmark's scoring, as in \cref{tab:spatial}.}
\label{fig:gal_vqa}
\end{figure*}

\Cref{fig:gal_clip,fig:gal_med,fig:gal_vqa} show one test item for every task of three pools: CLIP ViT-B/32 (\cref{tab:clip}), the medical and spatial pool on Qwen3-VL-2B (\cref{tab:medspatial}) and the natural-image pool on Qwen2-VL-7B (\cref{tab:spatial}). Each item is one that \method{} and the task's expert answer correctly and that most of the other merges get wrong; among those, we picked items that read well at this size. The tables give the rates.

On eight of the nine medical and spatial questions, the merges that miss all pick the same wrong option. On the natural-image pool, the misses include a shortened book title (TextVQA), another book genre (OCR-VQA) and a box around the shirt instead of the person (RefCOCO).

\Cref{tab:llm_examples} shows, for each of the eight generation tasks of the language-model pool at $K = 8$, a test prompt on which \method{} scores above every other merge, with each answer's ROUGE-L; \cref{tab:llm} gives the rates. On seven of the eight, \method{} gives the expert's answer word for word. The other merges lose the task's output format: all four repeat the review as its title on Amazon, TIES and WTA repeat the input on XLSum, and on CommonGen all four write a full sentence where the expert writes a caption. NLI answers are a class label, so we show none.

\input{tables/supp_llm_examples}

%% file: tables/supp_setup.tex
\begin{table*}[t]
\centering\small
\caption{\textbf{Setup.} Every adapter is a LoRA of rank 16. Calibration: unlabeled inputs per holder for the Grams. Pool: unlabeled inputs at the merger for the label-free pick (Gemma-4-31B: validation questions). Experts: mean raw accuracy of the experts on test; on the sites row, the range over the ten site adapters.}
\label{tab:setup}
\begin{tabular}{lrrrrr}
\toprule
Base model & Tasks & K & Test items & Experts (raw \%) & Unlabeled pool \\
\midrule
CLIP ViT-B/32 & Image classification, 8 tasks & 8 & 69,947 & 84.0 & 150 / task \\
CLIP ViT-L/14 & Image classification, 8 tasks & 8 & 69,947 & 92.4 & 150 / task \\
Qwen3-VL-2B & Medical VQA, 8 modalities & 8 & 2,400 & 99.5 & 250 / task \\
InternVL2-1B & Medical VQA, 8 modalities & 8 & 2,400 & 99.6 & 250 / task \\
Qwen3-VL-4B & Medical VQA, 8 modalities & 8 & 2,400 & 99.5 & 50 / task \\
Gemma-4-31B-it & Medical cls. + VQA, 8 tasks & 8 & 16,515 & 71.6 & 100 / task (val) \\
Qwen3-VL-2B (sites) & Dermoscopy VQA, 10 sites & 10 & 1,306 & 78.6--90.0 & 1,285 in all \\
\bottomrule
\end{tabular}
\end{table*}

%% file: tables/supp_judge_prompt.tex
\newcommand{\jbul}{\par\hangindent=1.6em\hangafter=1\noindent\hspace*{0.6em}\textbullet~}
\newcommand{\jhead}[1]{\par\smallskip\noindent\textbf{#1}\par}
\begin{table}[!htb]
\caption{\textbf{The LLM judge's prompt for the Gemma-4-31B VQA tasks}, verbatim from~\cite{zhao2026rgicl} (its Supplementary Fig.~3). Each request gives the judge the question, the reference answer and the merged model's answer, without the image. An answer is correct when \texttt{semantic\_accuracy} is 100, the endpoint of~\cite{zhao2026rgicl}; the judge is GPT-6 Luna in place of their GPT-5.4-mini (Supp.~A).}
\label{tab:judge}
\scriptsize
\begin{tcolorbox}[title=Medical VQA LLM-judge prompt, colback=white, colframe=black!60, fonttitle=\footnotesize\bfseries]
\raggedright
\textbf{System:} You are an impartial evaluator for a medical visual question answering benchmark.\medskip\\
\textbf{User:}
\jhead{Inputs:}
\jbul question: the question asked about a medical image
\jbul reference\_answer: the ground-truth/reference answer
\jbul predicted\_answer: the model's answer
\par\smallskip\noindent Evaluate the predicted answer with respect to the reference answer. Use the reference answer as the source of truth. You may use medical knowledge only to recognize harmless synonyms, abbreviations, spelling variants, and equivalent phrasing.
\jhead{Score four dimensions from 0 to 100:}
\jhead{1. semantic\_accuracy}
\noindent How well the prediction matches the meaning of the reference answer.
\jbul 100: exact semantic match; fully equivalent to the reference answer
\jbul 80: correct with only minor wording differences or minor omissions
\jbul 60: partially correct but missing important information
\jbul 40: weakly related but mostly incorrect
\jbul 0: incorrect, unrelated, contradictory, or non-answer
\jhead{2. completeness}
\noindent How fully the prediction covers the required information in the reference answer.
\jbul 100: covers all required concepts
\jbul 80: covers the main concept with only minor omissions
\jbul 60: covers some required concepts but misses important elements
\jbul 40: covers only a small part of the required answer
\jbul 0: does not cover the required information
\jhead{3. factuality}
\noindent Whether the prediction avoids unsupported or contradictory claims relative to the reference answer.
\jbul 100: no unsupported or contradictory claims
\jbul 80: mostly factual, with only harmless extra wording
\jbul 60: contains minor unsupported details that do not change the main answer
\jbul 40: contains notable unsupported or potentially misleading details
\jbul 0: contains contradictory, hallucinated, or clinically misleading claims
\jhead{4. conciseness}
\noindent Whether the prediction answers the question directly without unnecessary extra explanation.
\jbul 100: direct and concise
\jbul 80: mostly concise with minor extra wording
\jbul 60: somewhat verbose but still focused
\jbul 40: overly verbose, hedging, or includes unnecessary explanation
\jbul 0: rambling, evasive, or dominated by irrelevant content
\jhead{Return only valid JSON in the following format:}
\noindent\texttt{\{}\\
\hspace*{1em}\texttt{"semantic\_accuracy": <0-100>,}\\
\hspace*{1em}\texttt{"completeness": <0-100>,}\\
\hspace*{1em}\texttt{"factuality": <0-100>,}\\
\hspace*{1em}\texttt{"conciseness": <0-100>,}\\
\hspace*{1em}\texttt{"rationale": "<brief explanation>"}\\
\texttt{\}}
\end{tcolorbox}
\end{table}

%% file: tables/supp_spectral.tex
\begin{table}[t]
\centering\footnotesize\setlength{\tabcolsep}{3pt}
\caption{\textbf{\method{} leads the spectral and LOT merges on ViT-B/32.} Normalized test accuracy (\%) on our adapters and splits. Labeled: scale picked on labeled validation data; label-free: picked by agreement on the unlabeled pool. For LOT, agreement picks the same scale as the labels, $\lambda = 0.8$. Bold: best.}
\label{tab:spectral}
\begin{tabular}{llr}
\toprule
Method & Scale pick & ViT-B/32 \\
\midrule
LOT~\cite{sun2025lot} & $\lambda = 1$ & 72.4 \\
LOT~\cite{sun2025lot} & labeled & 76.3 \\
Core Space~\cite{panariello2025core} & labeled & 76.3 \\
Iso-C~\cite{marczak2025isoc} & labeled & 72.0 \\
Iso-CTS~\cite{marczak2025isoc} & labeled & 68.5 \\
\midrule
\oursrow \textbf{\method{} (Ours)} & label-free & \textbf{79.8} \\
\bottomrule
\end{tabular}
\end{table}

%% file: tables/supp_curves.tex
\begin{table*}[t]
\centering\small\setlength{\tabcolsep}{4pt}
\caption{\textbf{The label-free pick is the most accurate scale on every pool: the numbers behind Figure~\ref{fig:scale}.} \method{} (RegMean on base-model Grams) at every scored scale $c$: agreement with the eight adapters on the unlabeled pool and accuracy, normalized to the experts on test (Gemma-4-31B: raw accuracy on 100 validation questions per task). Bold: the label-free pick $c^\star$ (ties go to the smaller $c$). Every VLM scale is scored in fp32, on pool and test.}
\label{tab:curves}
\begin{tabular}{llrrrrrrrrrrrr}
\toprule
Pool & Measure & $0.025$ & $0.05$ & $0.075$ & $0.1$ & $1/8$ & $0.15$ & $0.2$ & $0.25$ & $0.3$ & $0.35$ & $0.4$ & $0.5$ \\
\midrule
CLIP ViT-B/32 & agreement & 56.8 & 63.5 & 68.3 & \textbf{70.8} & 69.6 & 64.8 & 48.1 & 35.6 & 26.7 & 19.0 & 10.8 & 6.8 \\
 & accuracy & 66.6 & 72.8 & 77.4 & \textbf{79.8} & 76.4 & 71.3 & 54.1 & 40.1 & 30.1 & 21.2 & 12.6 & 7.2 \\
\midrule
CLIP ViT-L/14 & agreement & 74.3 & 78.9 & \textbf{80.6} & 78.8 & 76.1 & 72.1 & 59.9 & 44.5 & 32.4 & 23.8 & 15.8 & 9.0 \\
 & accuracy & 76.5 & 80.7 & \textbf{81.9} & 80.5 & 78.1 & 73.9 & 61.5 & 46.8 & 34.3 & 24.0 & 16.8 & 9.5 \\
\midrule
Qwen3-VL-2B & agreement & 44.3 & 75.7 & 91.7 & 96.1 & 97.3 & 97.8 & \textbf{98.0} & 98.0 & 97.7 & 95.7 & 91.5 & 39.1 \\
 & accuracy & 44.4 & 73.6 & 91.3 & 96.3 & 97.5 & 98.2 & \textbf{98.6} & 98.4 & 98.2 & 96.1 & 91.5 & 37.8 \\
\midrule
InternVL2-1B & agreement & 63.7 & 71.9 & 80.7 & 87.9 & 92.3 & 94.8 & \textbf{95.8} & 93.4 & 85.5 & 68.1 & 52.5 & 36.0 \\
 & accuracy & 62.6 & 71.6 & 80.7 & 87.4 & 91.8 & 94.5 & \textbf{95.9} & 93.8 & 86.2 & 70.6 & 51.6 & 34.5 \\
\midrule
Gemma-4-31B (val.) & agreement & 61.2 & 68.3 & 74.4 & 78.2 & 81.1 & \textbf{83.5} & 83.5 & 82.0 & 77.3 & 66.5 & 51.2 & 39.9 \\
 & accuracy & 56.8 & 61.3 & 64.5 & 65.6 & 67.1 & \textbf{67.9} & 67.8 & 67.5 & 66.1 & 62.5 & 58.1 & 42.1 \\
\bottomrule
\end{tabular}
\end{table*}

%% file: tables/supp_experts.tex
\begin{table}[t]
\centering\footnotesize\setlength{\tabcolsep}{2.4pt}
\caption{\textbf{The label-free pick improves RegMean on its own Grams on every pool.} Accuracy of RegMean's merge on the experts' own Grams at another scale, in points gained over $c = 1/K$; pools and metric as in \cref{tab:ablate}. Our pick: the label-free pick of Eq.~(\ref{eq:pick}) applied to the experts' Grams (CLIP: $c = 0.1$ on both backbones; VLMs: 0.2; Gemma-4-31B: 0.15; sites: 0.15 on both splits). Bold: best per column.}
\label{tab:experts}
\begin{tabular}{lrrrrrrr}
\toprule
Scale & B/32 & L/14 & Qwen & IVL & Gemma & Sites 1 & Sites 2 \\
\midrule
Fixed $c = 0.2$ & $-38.7$ & $-27.6$ & $\mathbf{+0.9}$ & $\mathbf{+5.0}$ & $-21.2$ & $+0.4$ & $-38.4$ \\
Our pick & $\mathbf{+3.9}$ & $\mathbf{+2.7}$ & $\mathbf{+0.9}$ & $\mathbf{+5.0}$ & $\mathbf{+1.7}$ & $\mathbf{+1.8}$ & $\mathbf{+1.9}$ \\
\bottomrule
\end{tabular}
\end{table}

%% file: tables/supp_paired.tex
\begin{table*}[t]
\centering\small
\caption{\textbf{\method{}'s leads exceed twice their standard errors.} Paired differences on test (points; paired bootstrap SE over test items within each task or source, 2,000 draws; positive favors the first). Resampling whole images instead of questions changes the SEs by at most 0.02 on every pool but Gemma-4-31B, which is resampled by question. Llama-3-8B: \method{}'s pick is $1/K$, so the first and last columns compare the same merges. --: not scored.}
\label{tab:paired}
\begin{tabular}{lrrr}
\toprule
Pool & \method{} $-$ RegMean & \method{} $-$ best learned merge & Base $-$ experts' Grams at $1/K$ \\
\midrule
CLIP ViT-B/32 & +8.8 (0.29) & +10.2 (0.26), AdaMerging & +5.4 (0.24) \\
CLIP ViT-L/14 & +3.7 (0.18) & +3.3 (0.19), AdaMerging & $-$0.1 (0.13) \\
Qwen3-VL-2B & +1.0 (0.29) & +5.2 (0.48), NSC & 0.0 (0.16) \\
InternVL2-1B & +4.7 (0.53) & +17.1 (0.75), NSC & +0.7 (0.26) \\
Qwen3-VL-4B & +0.9 (0.20) & +3.6 (0.41), Expert Merging & -- \\
Ten sites, split 2 & +3.8 (0.62) & +4.3 (0.63), AdaMerging or NSC & +0.6 (0.37) \\
Llama-3-8B, NLI & +0.8 (0.27) & +11.0 (0.45), NSC & +0.8 (0.27) \\
Gemma-4-31B & +0.8 (0.36) & +7.6 (0.52), AdaMerging & 0.0 (0.32) \\
\bottomrule
\end{tabular}
\end{table*}

%% file: tables/supp_spatial.tex
\begin{table*}[t]
\centering\small\setlength{\tabcolsep}{4pt}
\caption{\textbf{\method{} has the best average on natural-image VQA and keeps the spatial expert (Qwen2-VL-7B, five rank-8 LoRA experts).} Four of OptMerge's experts~\cite{wei2026optmerge}, all but Geometry, and one trained on a 3DSRBench training split, merged into one LoRA. Absolute accuracy (\%) on test; RC: RefCOCO; Avg: mean over the eight benchmarks shown; 3DSR: 3DSRBench, CircularEval on 1,704 held-out questions. Fine-tuned: each expert on its own task. TA and TIES pick $\lambda$ on labeled validation data (0.7 and 1.5); \method{} picks its scale without labels ($c^\star = 0.25$). Bold: best merge in each column; underlined: second best.}
\label{tab:spatial}
\begin{tabular}{l|rr|r|rr|rrr|r|r}
\toprule
 & \multicolumn{2}{c|}{VQA} & Chart & \multicolumn{2}{c|}{OCR} & \multicolumn{3}{c|}{Grounding} & & Spatial \\
Method & VizWiz & GQA & ChartQA & TextVQA & OCRVQA & RC & RC+ & RCg & Avg & 3DSR \\
\midrule
Fine-tuned & 39.5 & 64.0 & 87.7 & 76.9 & 70.7 & 85.0 & 72.3 & 79.0 & 71.9 & 50.7 \\
\midrule
Weight averaging & 46.9 & \textbf{63.6} & 84.3 & \underline{79.6} & 66.3 & 81.3 & 64.7 & 80.3 & 70.9 & 42.1 \\
TA~\cite{ilharco2023editing} & \underline{51.1} & \textbf{63.6} & 86.0 & 75.3 & 66.0 & 76.3 & 68.0 & 83.0 & 71.2 & 49.9 \\
TIES~\cite{yadav2023ties} & \textbf{51.9} & 62.0 & 87.0 & 74.0 & 66.7 & 81.0 & 66.7 & 83.0 & 71.5 & \underline{50.5} \\
RegMean~\cite{jin2023regmean} & 44.9 & \underline{62.8} & \underline{87.3} & \textbf{81.3} & \underline{69.7} & \underline{84.7} & \underline{70.0} & \underline{84.3} & \underline{73.1} & 49.5 \\
\oursrow \textbf{\method{} (Ours)} & 47.6 & 61.2 & \textbf{88.3} & \textbf{81.3} & \textbf{70.0} & \textbf{85.0} & \textbf{73.0} & \textbf{85.3} & \textbf{74.0} & \textbf{50.9} \\
\bottomrule
\end{tabular}
\end{table*}

%% file: tables/supp_medspatial.tex
\begin{table*}[t]
\centering\small
\caption{\textbf{\method{} has the best average with medical and spatial experts in one pool (Qwen3-VL-2B, nine rank-16 LoRA experts).} MergeMedBench's eight modality experts~\cite{mou2026mergemedbench} and one trained on the 3DSRBench training split of \cref{supp:spatial}, merged into one LoRA. Fine-tuned: each expert on its own task, absolute accuracy (\%); merged models are normalized to it (\%). Med.: mean over the eight modalities (300 test questions each, as in \cref{tab:medvlm}); 3DSR: 3DSRBench, CircularEval on 1,704 held-out questions; Avg: mean over all nine tasks. All merges are label-free: TA and \method{} pick their scale by \cref{eq:pick} ($\lambda = 0.3$ and $c^\star = 0.15$), RegMean and weight averaging use $1/K$, TIES uses MergeMedBench's setting. Bold: best merge in each column; underlined: second best.}
\label{tab:medspatial}
\begin{tabular}{l|rrrrrrrr|r|r|r}
\toprule
Method & CT & MRI & X-ray & US & Derm. & Fundus & OCT & Micro. & Med. & 3DSR & Avg \\
\midrule
Fine-tuned & 100.0 & 99.7 & 98.7 & 100.0 & 98.0 & 100.0 & 100.0 & 99.7 & 99.5 & 48.1 & 93.8 \\
\midrule
Weight averaging & 71.3 & 89.3 & 88.9 & 89.0 & 83.0 & 93.0 & 75.7 & 85.3 & 84.4 & 82.8 & 84.2 \\
TA~\cite{ilharco2023editing} & 91.7 & 97.0 & 93.6 & 96.7 & 88.4 & 94.0 & 91.3 & 86.6 & 92.4 & 95.7 & 92.8 \\
TIES~\cite{yadav2023ties} & 95.0 & \underline{98.0} & 94.6 & 92.7 & 88.1 & 94.3 & 89.7 & 89.3 & 92.7 & \underline{101.7} & 93.7 \\
RegMean~\cite{jin2023regmean} & \underline{98.0} & \textbf{98.7} & \textbf{97.6} & \underline{99.3} & \underline{93.2} & \underline{98.3} & \underline{95.7} & \underline{95.7} & \underline{97.1} & 100.1 & \underline{97.4} \\
\oursrow \textbf{\method{} (Ours)} & \textbf{98.7} & \textbf{98.7} & \underline{97.0} & \textbf{99.7} & \textbf{93.9} & \textbf{99.0} & \textbf{98.7} & \textbf{96.3} & \textbf{97.7} & \textbf{102.2} & \textbf{98.2} \\
\bottomrule
\end{tabular}
\end{table*}

%% file: tables/supp_sensitivity.tex
\begin{table*}[t]
\centering\small\setlength{\tabcolsep}{4pt}
\caption{\textbf{\method{} needs few unlabeled inputs.} Left: base-model Grams from $n$ calibration questions per task on InternVL2-1B, merged at $c = 0.2$; normalized accuracy on 150 test questions per task, mean and range over three random draws of the questions. Right: the label-free pick on random subsets of the ten sites' unlabeled pool (split~1, 24 candidates, 2,000 draws per size, each site's share of the pool kept); regret: test accuracy of the best candidate minus that of the pick (points); last row: share of draws whose pick scores above the best single site.}
\label{tab:sensitivity}
\begin{tabular}{lrrrrrrr}
\toprule
\multicolumn{8}{l}{\textit{Calibration questions per task (InternVL2-1B)}} \\
$n$ & 4 & 8 & 16 & 32 & 64 & 128 & 256 \\
\midrule
Mean & 94.3 & 95.3 & 95.5 & 96.1 & 96.1 & 96.0 & 96.1 \\
Min & 92.3 & 93.7 & 94.4 & 95.7 & 95.9 & 95.7 & 95.9 \\
Max & 95.6 & 96.7 & 96.4 & 96.3 & 96.1 & 96.2 & 96.2 \\
\bottomrule
\end{tabular}
\hspace{1.5em}
\begin{tabular}{lrrrrr}
\toprule
\multicolumn{6}{l}{\textit{Pool questions for the pick (ten sites)}} \\
Questions & 53 & 102 & 257 & 513 & 1,285 \\
\midrule
Regret, mean & 1.44 & 0.62 & 0.27 & 0.13 & 0.00 \\
Above best site (\%) & 98.7 & 99.9 & 100 & 100 & 100 \\
\bottomrule
\end{tabular}
\end{table*}

%% file: tables/supp_llm_examples.tex
\begin{table*}[p]
\centering\scriptsize\setlength{\tabcolsep}{4pt}\renewcommand{\arraystretch}{0.92}
\caption{\textbf{Answers on each generation task.} One test prompt per task of the $K = 8$ Lots-of-LoRAs pool of \cref{tab:llm} (Mistral-7B), with each model's answer and its ROUGE-L ($\times 100$, best over the references). Each prompt is, among those with an input of at most 350 characters on which \method{} scores at least 50 and above every other merge, the one with \method{}'s largest lead over the best of them, among those where \method{} gives the expert's answer if any (all but WebNLG). The task's definition and worked examples precede the input and are not shown; answers are cut at the first newline, as scored.}
\label{tab:llm_examples}
\begin{tabular}{@{}p{0.1\textwidth}p{0.8\textwidth}r@{}}
\toprule
Model & Answer & R-L \\
\midrule
\multicolumn{3}{@{}p{\textwidth}@{}}{\textbf{XLSum title.} \textit{Input:} Dozens of Welsh victims of the contaminated blood scandal will be seeking compensation after a High Court ruling, an AM has said.} \\
\multicolumn{3}{@{}p{\textwidth}@{}}{\textit{Reference:} Contaminated blood: Welsh victims to seek compensation} \\
Expert & Contaminated blood victims to seek compensation & 92.3 \\
Averaging & Dozens of Welsh victims of contaminated blood scandal to seek compensation & 55.6 \\
TIES & Dozens of Welsh victims of the contaminated blood scandal will be seeking compensation after a High Court ruling, an AM has said. & 20.7 \\
WTA & Dozens of Welsh victims of the contaminated blood scandal will be seeking compensation after a High Court ruling, an AM has said. & 20.7 \\
RegMean & Compensation for Welsh victims of contaminated blood & 28.6 \\
\oursrow \method{} & Contaminated blood victims to seek compensation & 92.3 \\
\midrule
\multicolumn{3}{@{}p{\textwidth}@{}}{\textbf{Amazon review title.} \textit{Input:} Works as stated without issue so far} \\
\multicolumn{3}{@{}p{\textwidth}@{}}{\textit{Reference:} Five Stars} \\
Expert & Five Stars & 100.0 \\
Averaging & Works as stated without issue so far & 0.0 \\
TIES & Works as stated without issue so far & 0.0 \\
WTA & Works as stated without issue so far & 0.0 \\
RegMean & Works as stated without issue so far & 0.0 \\
\oursrow \method{} & Five Stars & 100.0 \\
\midrule
\multicolumn{3}{@{}p{\textwidth}@{}}{\textbf{WebNLG.} \textit{Input:} United\_Kingdom \textbar{} leaderName \textbar{} David\_Cameron
AIDS\_(journal) \textbar{} country \textbar{} United\_Kingdom} \\
\multicolumn{3}{@{}p{\textwidth}@{}}{\textit{Reference:} The AIDS Journal was published in the UK, where David Cameron was a leader. / AIDS (journal) is published in the UK where David Cameron is the leader. / AIDS journal is from the United Kingdom where the leader is David Cameron.} \\
Expert & The AIDS journal is from the United Kingdom where David Cameron is a leader. & 74.1 \\
Averaging & David Cameron is the leader of the United Kingdom and AIDS is a journal published in the United Kingdom. & 36.4 \\
TIES & David Cameron is the leader of the United Kingdom and AIDS (journal) is published in the United Kingdom. & 38.7 \\
WTA & David Cameron is the leader of the United Kingdom. & 45.5 \\
RegMean & David Cameron is the leader of the United Kingdom, where the journal AIDS is published. & 50.0 \\
\oursrow \method{} & The journal AIDS is published in the United Kingdom, where the leader is David Cameron. & 78.6 \\
\midrule
\multicolumn{3}{@{}p{\textwidth}@{}}{\textbf{NYC data-to-text.} \textit{Input:} name[xname], price[cheap], rating[5 out of 5], location[riverside]} \\
\multicolumn{3}{@{}p{\textwidth}@{}}{\textit{Reference:} Located on the riverside is xname.  It's known for having a cheap price range but a wonderful customer rating of 5 out of 5.} \\
Expert & xname is a cheap restaurant in the riverside area with a customer rating of 5 out of 5. & 51.2 \\
Averaging & xname is a cheap restaurant with a rating of 5 out of 5. It is located on the riverside. & 45.5 \\
TIES & xname is a cheap restaurant with a rating of 5 out of 5. It is located in riverside. & 46.5 \\
WTA & xname is a cheap restaurant located on the riverside. & 23.5 \\
RegMean & xname is a cheap restaurant in riverside with a 5 out of 5 customer rating. & 40.0 \\
\oursrow \method{} & xname is a cheap restaurant in the riverside area with a customer rating of 5 out of 5. & 51.2 \\
\midrule
\multicolumn{3}{@{}p{\textwidth}@{}}{\textbf{SQuAD question generation.} \textit{Input:} It is on Absecon Island, on the Atlantic coast. Atlantic City was incorporated on May 1, 1854, from portions of Egg Harbor Township and Galloway Township. The city borders Absecon, Brigantine, Pleasantville, Ventnor City and West Atlantic City.} \\
\multicolumn{3}{@{}p{\textwidth}@{}}{\textit{Reference:} What is the name of the island on which Atlantic City is located? / In what year was Atlantic City incorporated? / What are the names of the two townships that became part of Atlantic City when it was incorporated? / How many other towns share a border with Atlantic City? / Which coast is Atlantic City located on?} \\
Expert & What is the name of the island where Atlantic City is located? & 88.0 \\
Averaging & Where is Atlantic City located? & 66.7 \\
TIES & Where is Atlantic City? & 54.5 \\
WTA & What is Atlantic City, North Carolina? & 46.2 \\
RegMean & What is Atlantic City's location? & 46.2 \\
\oursrow \method{} & What is the name of the island where Atlantic City is located? & 88.0 \\
\midrule
\multicolumn{3}{@{}p{\textwidth}@{}}{\textbf{ParaNMT paraphrase.} \textit{Input:} move ! if you guys are cops , let 's see some badge .} \\
\multicolumn{3}{@{}p{\textwidth}@{}}{\textit{Reference:} if you 're cops , show the badges .} \\
Expert & if you 're cops , show me your badge . & 66.7 \\
Averaging & if you guys are cops, show me your badge ! & 50.0 \\
TIES & if you guys are cops, let's see some badges. & 47.1 \\
WTA & if you guys are cops, let's see some badge. & 35.3 \\
RegMean & if you guys are cops , show me your badge . & 50.0 \\
\oursrow \method{} & if you 're cops , show me your badge . & 66.7 \\
\midrule
\multicolumn{3}{@{}p{\textwidth}@{}}{\textbf{CommonGen.} \textit{Input:} cross\#hiker\#stream} \\
\multicolumn{3}{@{}p{\textwidth}@{}}{\textit{Reference:} Hikers are crossing a stream. / woman hiker crossing a frozen stream in rain forest / hikers cross a stream heading north towards us state} \\
Expert & hiker crossing a stream & 66.7 \\
Averaging & A hiker crosses a stream & 42.9 \\
TIES & A hiker crosses a stream. & 42.9 \\
WTA & A hiker crosses a stream & 42.9 \\
RegMean & A hiker crosses a stream. & 42.9 \\
\oursrow \method{} & hiker crossing a stream & 66.7 \\
\midrule
\multicolumn{3}{@{}p{\textwidth}@{}}{\textbf{Story ending.} \textit{Input:} Sentence1: Jeremy was going on his first fishing trip. Sentence2: Jeremy's dad took him out on the lake in a little rowboat. Sentence3: With his hook baited, Jeremy put his line in the water. Sentence4: It wasn't long before Jeremy got a bite.} \\
\multicolumn{3}{@{}p{\textwidth}@{}}{\textit{Reference:} He was so excited to catch the biggest fish he had ever seen.} \\
Expert & Jeremy was so excited to catch his first fish. & 54.5 \\
Averaging & Jeremy was so excited he couldn't wait to tell his friends. & 32.0 \\
TIES & Jeremy was so excited he couldn't wait to tell his friends. & 32.0 \\
WTA & Jeremy was happy. & 12.5 \\
RegMean & Jeremy was so excited he jumped up and down. & 36.4 \\
\oursrow \method{} & Jeremy was so excited to catch his first fish. & 54.5 \\
\bottomrule
\end{tabular}
\end{table*}